\documentclass[preprint,12pt]{elsarticle}

\let\today\relax
\makeatletter
\def\ps@pprintTitle{%
    \let\@oddhead\@empty
    \let\@evenhead\@empty
    \def\@oddfoot{\footnotesize\itshape
         {Preprint submitted} \hfill\today}%
    \let\@evenfoot\@oddfoot
    }
\makeatother

\usepackage{amssymb}

\usepackage{lineno}
\usepackage{hyperref}

\usepackage{pgfplots}
\usepackage{pgf, tikz}
 
\usetikzlibrary{arrows, automata}
\usetikzlibrary{shapes.multipart}
\usetikzlibrary{quotes,arrows.meta}%
\usetikzlibrary{decorations.pathreplacing}
\usetikzlibrary{positioning}
\usetikzlibrary{matrix}
\usetikzlibrary{backgrounds}
\usetikzlibrary{fit, patterns}
\usepackage[utf8]{inputenc}
\usepgfplotslibrary{groupplots,dateplot}
\usetikzlibrary{shapes.arrows}
\pgfplotsset{compat=newest}
\usepackage{graphicx}
\usepackage{graphbox}
\usepackage{color}
\usepackage{ifthen}
\usepackage{bm}
\usepackage{xurl}
\usepackage{adjustbox}
\usepackage{subcaption}
\usepackage{comment}
\usepackage{multirow}
\usepackage{xspace}
\usepackage{pgfgantt}

\usepackage{amssymb}
\usepackage{pdfpages}
\makeatletter
\newcommand\saveequation[2]{%
  \@namedef{equation@#1}{#2}%
}
\newcommand\useequation[1]{%
  \@nameuse{equation@#1}%
}
\makeatother
\RequirePackage{amsmath,bm}

\usepackage[figuresright]{rotating}

\usepackage{booktabs}

\usepackage[disable, colorinlistoftodos,textsize=small,textwidth=2.5cm]{todonotes}
\setuptodonotes{backgroundcolor=yellow!10!white}
\newcounter{todoListItems}

\usepackage[normalem]{ulem}

\def\methodname{DDCNet}

\saveequation{massenergy}{E = mc^{2}}
\saveequation{curlE}{\nabla \times \bm{E} = 0}
\newcommand{\filterNum}[1]{\ensuremath{M^{[#1]}}}
\newcommand{\filterStride}[1]{\ensuremath{s^{[#1]}}}
\newcommand{\filterDilation}[1]{\ensuremath{r^{[#1]}}}
\newcommand{\activationFunction}[1]{\ensuremath{g^{[#1]}}}
\newcommand{\featureMap}[1]{\ensuremath{\bm{A}^{[#1]}}}
\newcommand{\filter}[2]{\ensuremath{F^{[#1]}_{#2}}}
\saveequation{layerActivation}{\featureMap{\ell} = \activationFunction{\ell}(\bm{Z}^{[\ell]}) = \activationFunction{\ell}(\featureMap{\ell-1} * \filter{\ell}{})}

\includecomment{optionalQuestions}
\newcommand{\question}[1]{}
\begin{optionalQuestions}
	\renewcommand{\question}[1]{\noindent\textcolor{gray}{\newline  #1 \newline}}
\end{optionalQuestions}

\newcounter{myexample}[section]

\makeatletter
\DeclareRobustCommand\onedot{\futurelet\@let@token\@onedot}
\def\@onedot{\ifx\@let@token.\else.\null\fi\xspace}

\def\eg{\emph{e.g}\onedot} 
\def\ie{\emph{i.e}\onedot} 
 
\def\etc{\emph{etc}\onedot} 
 
\def\etal{\emph{et al}\onedot}
\makeatother

\newcommand{\tabhead}[1]{\textbf{#1}}
\providecommand{\decoRule}{\rule{.8\textwidth}{.4pt}} 

\def\minNeuronSize{1.29}%
\def\layerDistance{4.5}
\def\nodeDistance{1.3}
\tikzstyle{neuron} = [rectangle,fill=black!25,minimum size=\minNeuronSize cm,inner sep=0pt]
\tikzstyle{inputNeuronStyle} = [neuron, fill=green!50]
\tikzstyle{outputNeuronStyle}=[neuron, fill=red!50]
\tikzstyle{hiddenNeuronStyle}=[neuron, fill=blue!50]
\tikzstyle{zeroNeuronStyle}=[neuron, fill=gray!20]
\tikzstyle{annotatonStyle}=[text width=3cm,text centered, scale=1]
\tikzstyle{convInfoStyle}=[text width=4cm, text centered, rotate=0, scale=0.9]
\tikzstyle{moduleBox} = [rectangle, draw, inner sep=0pt, dashed, rounded corners,opacity=0.2,fit=#1]
\newcommand{\layerOneD}[3]{%
    \def\layerId{\tempa}
    \def\neuronNum{\tempb}
    \def\zeroPadNumTop{\tempc}
    \def\zeroPadNumBot{\tempd}
    \def\nodeContentM{\tempe} %
    \def\nodeContentSup{\tempf} %
    \def\nodeContentSub{\tempg} %
    \def\stx{\temph}
    \def\sty{\tempi}
    \def\annotationTop{#1}
    \def\annotationBot{#2}
    \def\stl{#3};
    \def\x{\stx}
    \def\y{\sty}
    \ifx\annotationTop\empty
        \pgfmathparse{\y+\nodeDistance}
        \edef\y{\pgfmathresult}
    \else
        \node[annotatonStyle] (\layerId-nt) at (\x, \y) {\annotationTop};
    \fi
    \ifnum\zeroPadNumTop>-1
        \pgfmathparse{\y-\nodeDistance}
        \edef\y{\pgfmathresult}
        \foreach \i in {0,...,\zeroPadNumTop}{
        	\path[xshift=0] 
    	    node[zeroNeuronStyle] (\layerId-zt-\i) at (\x,\y-\nodeDistance*\i) {$0$};
        	}
        \pgfmathparse{\y-\zeroPadNumTop*\nodeDistance}
        \edef\y{\pgfmathresult}
    \fi
    \pgfmathparse{\y-\nodeDistance}
    \edef\y{\pgfmathresult}
    \pgfmathparse{\neuronNum-1}
    \edef\neuronNum{\pgfmathresult}
    \foreach \i in {0,...,\neuronNum}{
    	\path[yshift=0] 
    	node[\stl] (\layerId-\i) at (\x,\y-\nodeDistance*\i) {$\nodeContentM^{\nodeContentSup}_{\i}$};
    }
    \pgfmathparse{\y-\neuronNum*\nodeDistance}
    \edef\y{\pgfmathresult}
    \ifnum \zeroPadNumBot>-1
        \pgfmathparse{\y-\nodeDistance}
        \edef\y{\pgfmathresult}
        \foreach \i in {0,...,\zeroPadNumBot}{
        	\path[yshift=0] 
        	node[zeroNeuronStyle] (\layerId-zb-\i) at (\x,\y-\nodeDistance*\i) {$0$};
        }
       \pgfmathparse{\y-\zeroPadNumBot*\nodeDistance}
        \edef\y{\pgfmathresult}
    \fi
    \ifx\annotationBot\empty
    \else
        \pgfmathparse{\y-\nodeDistance}
        \edef\y{\pgfmathresult}
        \node[annotatonStyle] (\layerId-nb) at (\x, \y) {\annotationBot};
    \fi
}
\newcommand{\dotlayer}[5]{%
    \def\layerId{#1}
    \def\stx{#2}
    \def\sty{#3}
    \def\notationy{#4}
    \def\stl{#5}
    \node[\stl, scale=2] (\layerId-dot) at (\stx, \sty) {$\dots$};
    \node[\stl] (\layerId-dot-n) at (\stx, \notationy) {\dots};
}%
\newcommand{\textOnlyLayer}[4]{%
    \def\layerId{#1}
	\def\stx{#2}
	\def\sty{#3}
	\def\txt{#4}
	\node[convInfoStyle] (\layerId-t) at (\stx, \sty) {\txt};
}%
\newcommand{\layerFlat}[7]{%
	\def\layerId{#1}
	\def\stx{#2}
	\def\sty{#3}
	\def\annotationTop{#4}
	\def\annotationInside{#5} %
	\def\annotationBot{#6}
	\def\stl{#7};
	\def\x{\stx}
	\def\y{\sty}
	\ifx\annotationTop\empty
	\pgfmathparse{\y+\nodeDistance}
	\edef\y{\pgfmathresult}
	\else
	\node[annotatonStyle] (\layerId-nt) at (\x, \y) {\annotationTop};
	\fi
	\pgfmathparse{\y-\nodeDistance}
	\edef\y{\pgfmathresult}
	\node (\layerId) at (\x,\y- 0.5*\minHeight) [\stl] {\rotatebox{90}{\annotationInside}};
	\pgfmathparse{\y-\minHeight}
	\edef\y{\pgfmathresult}
	\ifx\annotationBot\empty
	\else
	\pgfmathparse{\y-\nodeDistance}
	\edef\y{\pgfmathresult}
	\node[annotatonStyle] (\layerId-nb) at (\x, \y) {\annotationBot};
	\fi
}%

 \newcommand{\layerOneDErf}[3]{%
    \def\layerId{\tempa}
    \def\neuronNum{\tempb}
    \def\zeroPadNumTop{\tempc}
    \def\zeroPadNumBot{\tempd}
    \def\nodeContentA{\tempe} %
    \def\nodeContentB{\tempf} %
    \def\nodeContentC{\tempg} %
    \def\stx{\temph}
    \def\sty{\tempi}
    \def\annotationTop{#1}
    \def\annotationBot{#2}
    \def\stl{#3};
    \def\x{\stx}
    \def\y{\sty}
    \ifx\annotationTop\empty
        \pgfmathparse{\y+\nodeDistance}
        \edef\y{\pgfmathresult}
    \else
        \node[annotatonStyle] (\layerId-nt) at (\x, \y) {\annotationTop};
    \fi
    \ifnum\zeroPadNumTop>-1
        \pgfmathparse{\y-\nodeDistance}
        \edef\y{\pgfmathresult}
        \foreach \i in {0,...,\zeroPadNumTop}{
        	\path[xshift=0] 
    	    node[zeroNeuronStyle] (\layerId-zt-\i) at (\x,\y-\nodeDistance*\i) {$-$};
        	}
        \pgfmathparse{\y-\zeroPadNumTop*\nodeDistance}
        \edef\y{\pgfmathresult}
    \fi
    \pgfmathparse{\y-\nodeDistance}
    \edef\y{\pgfmathresult}
    \pgfmathparse{\neuronNum-1}
    \edef\neuronNum{\pgfmathresult}
    \foreach \i in {0,...,\neuronNum}{
        \ifnum \layerId=2
            \def\val{0}
            \ifnum \i=\nodeContentB
                \def\val{1}
            \fi
            \path[yshift=0] node[\stl] (\layerId-\i) at (\x,\y-\nodeDistance*\i) {$\frac{\partial \nodeContentA}{\partial \nodeContentC_{\i}}{=}\val$};
        \else
        	\path[yshift=0] 
        	node[\stl] (\layerId-\i) at (\x,\y-\nodeDistance*\i) {$\frac{\partial \nodeContentA_\nodeContentB}{\partial \nodeContentC_{\i}}$};
        \fi
    }
    \pgfmathparse{\y-\neuronNum*\nodeDistance}
    \edef\y{\pgfmathresult}
    \ifnum \zeroPadNumBot>-1
        \pgfmathparse{\y-\nodeDistance}
        \edef\y{\pgfmathresult}
        \foreach \i in {0,...,\zeroPadNumBot}{
        	\path[yshift=0] 
        	node[zeroNeuronStyle] (\layerId-zb-\i) at (\x,\y-\nodeDistance*\i) {$-$};
        }
       \pgfmathparse{\y-\zeroPadNumBot*\nodeDistance}
        \edef\y{\pgfmathresult}
    \fi
    \ifx\annotationBot\empty
    \else
        \pgfmathparse{\y-\nodeDistance}
        \edef\y{\pgfmathresult}
        \node[annotatonStyle] (\layerId-nb) at (\x, \y) {\annotationBot};
    \fi
}

\makeatletter
\long\def\ifnodedefined#1#2#3{%
    \@ifundefined{pgf@sh@ns@#1}{#3}{#2}%
}
\makeatother

\tikzset{lossLableStyle/.style = {anchor=north west, inner sep=0, outer sep=0,fill opacity=.5,fill=white, text opacity=1, scale=0.5}}
\usepackage{tabularx}
\let\cite\cite

\journal{Image and Vision Computing}

\begin{document}
\newif\ifbone
\bonetrue 

\begin{frontmatter}



\title{DDCNet: Deep Dilated Convolutional Neural Network for Dense Prediction}


\author{Ali Salehi}
\author{Madhusudhanan Balasubramanian}

\address{Department of Electrical and Computer Engineering, The University of Memphis, Memphis TN 38152}

\begin{abstract}
Dense pixel matching problems such as optical flow and disparity estimation are among the most challenging tasks in computer vision. Recently, several deep learning methods designed for these problems have been successful. A sufficiently larger effective receptive field (ERF) and a higher resolution of spatial features within a network are essential for providing higher-resolution dense estimates. In this work, we present a systemic approach to design network architectures that can provide a larger receptive field while maintaining a higher spatial feature resolution. To achieve a larger ERF, we utilized dilated convolutional layers.  By aggressively increasing dilation rates in the deeper layers, we were able to achieve a sufficiently larger ERF with a significantly fewer number of trainable parameters. We used optical flow estimation problem as the primary benchmark to illustrate our network design strategy.  The benchmark results (Sintel, KITTI, and Middlebury) indicate that our compact networks can achieve comparable performance in the class of \textit{lightweight} networks.
\end{abstract}


\end{frontmatter}


\section{Introduction}
\label{sec:introduction}
Deep learning methods, especially convolutional neural networks (CNNs), have been successful and in some cases surpassed other classical methods for \textit{pixel-level prediction} problems. Semantic segmentation \cite{Long2015, liu2019auto}, next frame prediction \cite{Lotter2016, zhang2019flow}, depth estimation from a single image \cite{gur2019single, Eigen2014} and optical flow estimation \cite{sun2014quantitative} are among these problems which can be solved effectively using machine learning algorithms.

Current models for optical flow estimation generally adapt existing CNN architectures that are designed inherently for other computer vision tasks such as image classification or semantic segmentation \cite{dosovitskiy2015flownet,ilg2017flownet}. The optical flow problem, however, is substantially different from all these other computer vision tasks. Unlike image classification procedures which require a holistic, scale and shift-invariant approaches, flow estimation involves pixel-level coordinate and photometric transformations and require correspondence of spatial features.

Considering the spatiotemporal nature of the dense flow estimation problem, we have designed a \textit{receptive field}-guided network that takes two images (reference frame and its next frame) as input and generates a motion vector for every pixel in the reference image. Our systematic network design 1) has much fewer parameters when compared to other learning-based methods; 2) can estimate large motions while maintaining fine motion resolution; 3) does not suffer from the \textit{vanishing problem} of estimating optical flow in regions of a scene with smaller objects moving at a faster velocity and 4) can perform well in the presence of excessive occlusion. 

While dilated convolutions are often used to design networks with a larger receptive field such as for semantic segmentation \cite{yu2015multi, chen2018deeplab}, none of the existing designs, however, go beyond several consecutive layers of convolutions with higher dilation rates \cite{schuster2019sdc, hamaguchi2018effective, zhu2018learning}.  Increasing dilation rates by a factor of two in successive convolutional layers doubles the receptive field of the network, This, however, introduces severe \textit{gridding artifacts} which may limit the accuracy and resolution of the estimates for dense prediction problems \cite{wang2018smoothed}. One of our key findings from this research work is that increasing the dilation rate linearly (instead of an exponential increase) results in a faster rate of increase in size of the receptive field with minimal gridding artifacts and a desirable Gaussian ERF of the network.

The main contributions of this work are as follows:
\begin{itemize}
    \item A simpler design strategy to build compact networks for dense prediction problems such as optical flow estimation using standard network layers.
    \item A new design strategy to effectively increase the network receptive field with no or minimal reduction in feature resolution and to minimize gridding problem.
    \item We introduce two \textit{lightweight networks} called \textit{DDCNet}s that can serve as building blocks for constructing more complex networks.
\end{itemize}

\section{Related Work}
\label{sec:related}
While the primary focus of this work is on deep learning based methods for optical flow estimation, a comprehensive review of optical flow estimation methods is available elsewhere \cite{sun2014quantitative}.

\emph{FlowNet} is a pioneer supervised CNN-based framework developed for learning optical flow tasks from ground truth motion vectors \cite{Dosovitskiy2015}. FlowNet-Simple is designed as an encoder-decoder structure. Flownet-Correlation is a variation of FlowNet-Simple that uses a custom layer called correlation layer to explicitly match feature maps extracted from each image in a sequence. Both methods lack the ability to recover high-resolution features needed to accurately estimate optical flow and clear motion boundaries. These methods use a variational approach as the last optional refinement to improve the estimation. 

\emph{Flownet2} improved upon the FlowNets by stacking multiple networks, adding a small-displacement estimation network, and by scheduling the order of presenting the training data to the network \cite{Ilg2017}. In a stacked network, higher-level blocks take the estimated flows from the previous blocks alongside the copies of input images as their input. Some of these FlowNet-Simple blocks were trained on specific datasets to achieve higher accuracy in estimating smaller displacements. Their experiments show that using intermediate flow estimates to warp one of the images and using the difference between the warped image and the reference image further improves the performance of FlowNet2. \emph{DispNet} extends this idea of supervised learning for optical flow estimation to disparity and scene flow estimation \cite{mayer2016large}.

Other methods such as a 3D Voxel2Voxel CNN \cite{tran2016deep}, unsupervised learning \cite{ahmadi2016unsupervised, jason2016back}, and a pyramidal approach \cite{ranjan2017optical} have been developed based on FlowNet. Most of these methods follow an encoder-decoder architecture \cite{kendall2015bayesian} with convolutional layers are stacked in the encoder section to extract high-level features from input data. Convolutional strides more than one in the downsampling convolutional layers in the encoder section are used to increase the receptive field of the network. The decoder section attempts to recover the feature resolution using deconvolutional (or fractionally-strided convolutions) layers \cite{zeiler2010deconvolutional} or upsampling layers followed by a convolutional layer. 


\emph{Teney \etal} \cite{teney2016learning} designed a CNN for optical flow estimation inspired by the classical spatiotemporal motion-energy filters of Heeger, \etal \cite{Heeger1988}. Their design requires much fewer training samples compared to FlowNet. They achieve this by enforcing different constraints on learned filters at different layers of the network.

\emph{SpyNet} is much smaller than FlowNet models and performs better on some standard benchmarks but also suffers from issues associated with classical spatial-pyramid approaches \cite{ranjan2017optical}.  \emph{PWC-Net} and \emph{LiteFlowNet} models use feature warping instead of image warping along with correlation layers as in FlowNet-Correlation \cite{sun2018pwc}.  LiteFlowNet is among the best-performing methods in the lightweight deep learning category. It uses convolution layers with strides of two to build a pyramid of feature maps. A descriptor matching approach is used to obtain a coarser estimate of optical flow using a custom layer similar to the correlation layer used in the FlowNet-Correlation network. Feature maps are warped (as in PWC-Net) from the finer levels of the pyramid using coarser flow estimates and a descriptor matching is applied to obtain a finer flow estimate \cite{hui2018liteflownet}. The feature warping steps improve the efficiency of descriptor matching. However, errors introduced in the coarser levels are propagated to the finer levels in the presence of occlusion and homogeneous regions. \emph{LiteFlowNet2} and \emph{LiteFlowNet3} are designed to address the issues associated with pyramidal error propagation. These models are faster and more accurate than the original LiteFlowNet \cite{hui2019lightweight, hui2020liteflownet3}.  

Spatial pyramid approach has two other major issues namely \emph{vanishing problem} due to smaller objects moving at a faster velocity in the scene, and \emph{ghosting artifact} when warping images (as in FlowNet2 and SpyNet) using coarser-level flow estimates in a scene; for example, ghosting artifact in a warped image due to motion of an occluding object over a stationary background \cite{janai2018unsupervised}.  In contrast to the LiteFlowNet and SpyNet, our model does not utilize a coarse-to-fine approach and therefore does not suffer from \emph{ghosting artifacts} \cite{janai2018unsupervised}. Our approach is to design networks with appropriate receptive fields and allow the network to learn network parameters necessary for estimating larger motions.

All three LiteFlowNets have complicated network structures. The training procedure of the LiteFlowNet3 is also complicated as they introduce extra network modules after initial training, adjust learning rates for these modules and proceed with training the whole network.
    
Several \emph{unsupervised} networks are also designed to tackle the optical flow problem. The matching task that is done by the network, depends on the structure of the network and is mostly similar to unsupervised models. The major difference between supervised and unsupervised networks is in the choice of the loss function. In general, the loss function of these methods is based on classical formulations of flow estimation, brightness constancy, and smoothness assumptions.  To train an unsupervised method, there is no need for ground-truth flow vectors, only two input images are sufficient. In such methods, the cost function for training is defined without requiring a knowledge of the ground truth motion in the input sequence. Typically a backward warping procedure is used to quantify the accuracy of the network flow estimated indirectly by comparing the first frame with the second frame warped using the estimated optical flow.

\emph{Dense Spatial Transform Flow} (DSTFlow) is among the earlier unsupervised networks and is based on the Spatial Transformer Network (STN) \cite{ren2017unsupervised}. DSTFlow learns optical flow by directly minimizing photometric constancy. The cost function is designed based on variational methods. USCNN \cite{ahmadi2016unsupervised}, DSTFlow \cite{ren2017unsupervised}, UnFlow \cite{meister2017unflow}, \etc are among the methods in this category. \emph{Back to Basics} is an end-to-end and unsupervised method based on variational loss \cite{jason2016back}. This method uses the FlowNet-Simple network structure but with an unsupervised loss function. \emph{USCNN} is also among the early unsupervised CNNs for optical flow estimation. Ahmadi \etal utilized the brightness constancy term of Horn-Schunck to train this network \cite{ahmadi2016unsupervised}.

Several other methods such as \cite{lai2017semi} can be categorized as semi-supervised methods. These methods have more complex cost functions based on several terms from various supervised and unsupervised methods. 

    
        
\section{Deep Dilated Convolutional Neural Network (DDCNet)}
\label{sec:DDCNet}
\textit{Receptive field} (RF) of a neuron in the output layer represents the spatial extent of its visibility and access to a neighborhood of pixels in the input image sequence.  \textit{Effective receptive field} (ERF) represents the actual RF of a trained network \cite{luo2017understanding}.  Therefore, to detect presence of large motion and to estimate large flow vectors, each neuron in the output layer should have a sufficiently large ERF. 

ERF of a network depends on two main factors namely the network architecture (arrangement and type of the layers) and the learned network weights.  Size, stride length, and activation functions of filters, network depth, and use of other types of layers contribute to the final ERF of output units in a CNN. Even in fully connected networks, the learned weights can suppress the influence of selected input neurons on the network response. Therefore, we utilized ERF instead of RF as a guiding criterion for designing our networks.

We defined the spatial extent of ERF based on the access and visibility of the neurons in one of the input channels with respect to the central neuron in the $u$ channel of the output layer. We numerically estimated the ERF of a network as the partial derivative of output neuronal responses with respect to each of the input neurons \cite{luo2017understanding}. For general analysis, we focus on the central unit of just one output channel ($u$ channel) in the network. Therefore, this ERF-map represents weighted contributions of each of the input neurons (image pixels) on the response of the central unit in the $u$ channel of the output layer as a gradient measure (defined similarly for the $v$ channel).

\def\sharedCaptionA{3D surface}
\def\sharedCaptionB{2D view of map}
\def\sharedCaptionC{Central row (with its FWHM)}
\edef\mainLabel{fig:ErfSampleNetOne}
\begin{figure}[htbp]
    \centering
    \subcaptionbox{\sharedCaptionA \label{\mainLabel:a}}[0.32\linewidth]{\includegraphics[trim={3.5cm 2.5cm 3.5cm 0},clip, width=\linewidth]{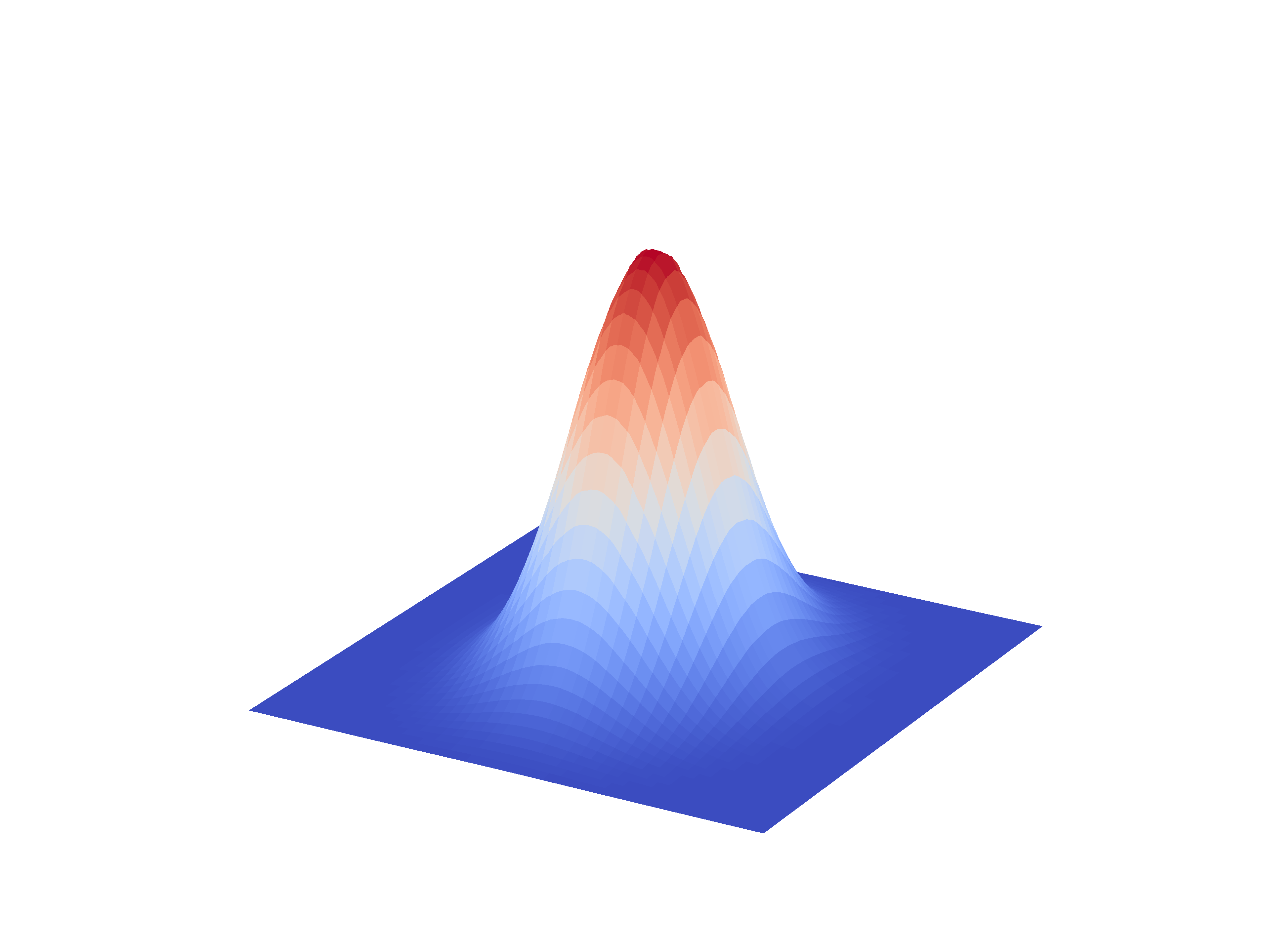}}
    \subcaptionbox{\sharedCaptionB \label{\mainLabel:b}}[0.32\linewidth]{\includegraphics[width=0.85\linewidth]{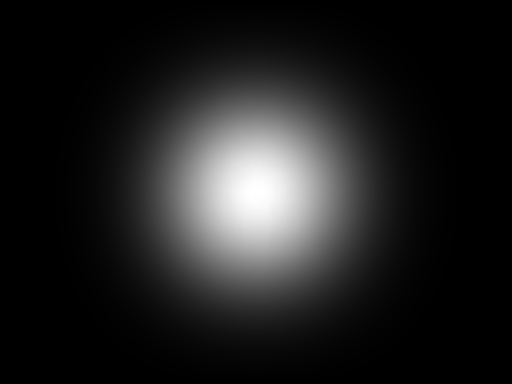}}
    \subcaptionbox{\sharedCaptionC
    \label{\mainLabel:c}}[0.33\linewidth]{\scalebox{
    0.5}{\input{Figures/ERF/B0/1.tex}}}
    \decoRule
    \caption[Receptive field of B0]{Effective Receptive Field of central output unit from DDCNet-B0.  All kernels of the network were initialized with a constant weight. \subref{\mainLabel:a} shows surface plot of the ERF map,  \subref{\mainLabel:b} shows the ERF as gray-scale image. \subref{\mainLabel:c} shows the 1-D plot of the central row in the ERF map and its FWHM.} 
    \label{\mainLabel}
\end{figure}

ERF maps can be visualized as 3D surfaces or gray-scale images as in Figure~ \ref{fig:ErfSampleNetOne} for visual assessment of network ERF and for selecting candidate networks.  For quantitative evaluation, we derived a measure of Full Width at Half Maximum (FWHM\footnote{Full width at half maximum (FWHM) of distribution represents the width or spread of the distribution at half of the maximum frequency or density value.}) from a Gaussian model of ERF weights along the center row of the ERF map (\ie the row corresponding to the target output unit) as in Figure~\ref{fig:ErfSampleNetOne:c}.

\subsection{Extent and Shape of the Network ERF}
    Candidate optical flow networks should be able to find matching pixels in the input sequence and therefore requires a sufficiently larger ERF extent. More precisely, the network ERF should have coverage and visibility of matched pixels corresponding to the \textit{largest motions in the data}. Therefore, our first design criterion is that the \emph{FWHM of ERF of candidate networks should be larger than the flow magnitudes for majority of the pixel locations}.
    
    ERF extent can also be increased by adding more convolutional layers, however, it increases linearly with the network depth as shown in Figure \ref{fig:ErfRegularVsDilatedNet:Regular}.  Deeper networks not only have more parameters to learn but also are more challenging to train due to the vanishing gradient problem \cite{huang2016deep}.  Choosing larger filter kernels in the first network layer may provide a larger ERF.  It is less effective in CNNs as it drastically increases the learning parameters \cite{mishkin2017systematic}.
    
    \textit{Pooling layers} allows a network to be spatially-invariant and allow the ERF extent to grow exponentially as a function of number of pooling layers as in FlowNet \cite{ilg2017flownet, dosovitskiy2015flownet} and LiteFlowNet \cite{hui2018liteflownet, hui2019lightweight, hui2020liteflownet3}. They are useful for classification tasks but undesirable for tasks such as optical flow where higher spatial resolution of features is likely desired.
    
    Unlike the pooling layers, dilated layers \cite{yu2015multi} do not cause loss of spatial resolution.  We illustrate the effectiveness of dilated convolutional layers in Figure \ref{fig:ErfRegularVsDilatedNet} by comparing the ERF of a network with 30 convolutional layers (Figure~\ref{fig:ErfRegularVsDilatedNet:Regular}) to a a network with 30 dilated-convolutional layers with increasing dilation rates \ref{fig:ErfRegularVsDilatedNet:Dilated}.  In both the networks, all hyper-parameters (\eg number of layers, size of filters, activation functions, \etc) were kept the same. It can be observed that a standard convolutional network would require hundreds of convolutional layers to achieve the ERF extent of the dilated-network.  Therefore, increasing the depth is not an effective approach to achieving a larger ERF.
    
    \begin{figure}[htbp]
        \centering
        \edef\mainLabel{fig:ErfRegularVsDilatedNet}
        \subcaptionbox{With incremental dilation rates \label{\mainLabel:Dilated}}{\includegraphics[width=0.4\linewidth]{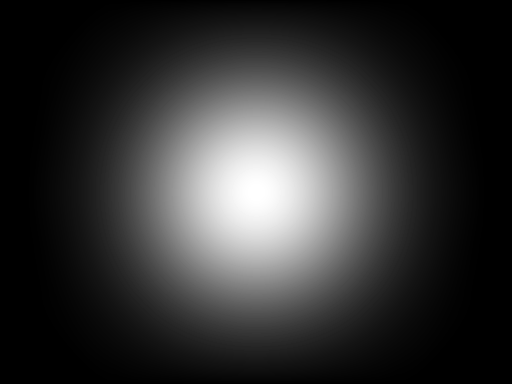}}
        \subcaptionbox{Dilation rates of 1 for all layers \label{\mainLabel:Regular}}{\includegraphics[width=0.4\linewidth]{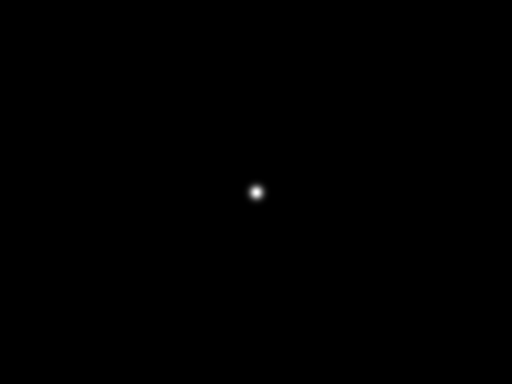}}
        \decoRule
        \caption[ERF comparison between dilated and regular CNNs]{ERF of dilated CNN vs classical CNN with 30 layers. \subref{\mainLabel:Dilated}: ERF when dilation rates increased from 1 through 30 in steps of 1. \subref{\mainLabel:Regular}: ERF when dilation rate kept at 1 for all 30 layers \ie no dilation.}
        \label{\mainLabel}
    \end{figure}
    
    \emph{ERF shape} is also an important determinant of network performance. While several approaches are available to achieve a bigger ERF, not all of them provide an \emph{optimal} receptive field leading to the highest network accuracy. For example, a larger ERF with a much shallower network can be achieved by rapidly increasing the dilation rates between subsequent dilated convolution layers (\eg doubling dilation rates). This, however, introduces undesirable \textit{ERF gridding artificats}.  Figure \ref{fig:ErfBadExamples} shows a few examples of undesirable ERFs.
    
    \begin{figure}[htbp]
        \centering
        \edef\mainLabel{fig:ErfBadExamples}
        \subcaptionbox{ERF with gaps \label{\mainLabel:a}}{\includegraphics[width=0.24\linewidth]{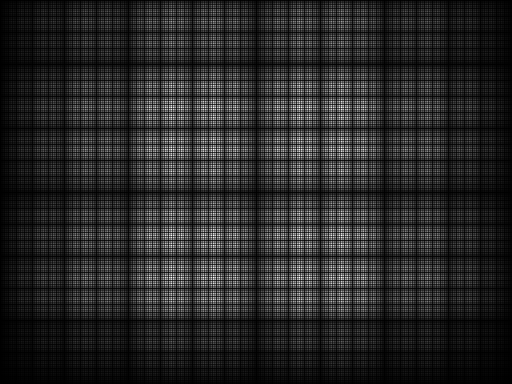}}
        \subcaptionbox{Gridded ERF \label{\mainLabel:b}}{\includegraphics[width=0.24\linewidth]{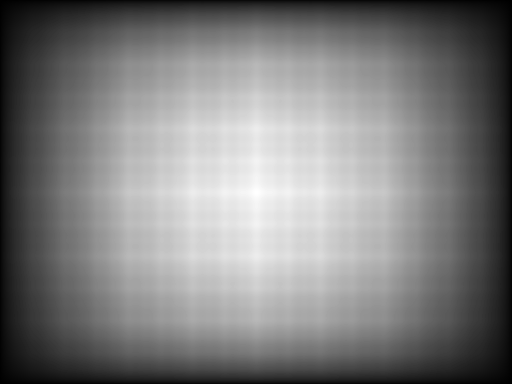}}
         \subcaptionbox{Non Gaussian ERF \label{\mainLabel:c}}{\includegraphics[width=0.24\linewidth]{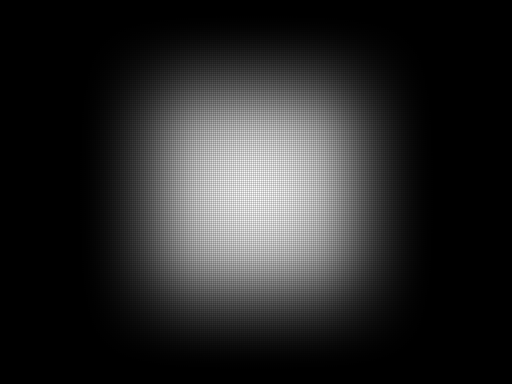}}
        \decoRule
        \caption[Samples of bad ERF shape]{Examples of networks with undesirable ERF shapes.}
        \label{\mainLabel}
    \end{figure}
    
    In general, when the temporal sampling rate is high, it is expected that the majority of the pixels have smaller flow magnitudes.  Even when the frame rate is high, several locations may exhibit larger flow magnitudes due to relative motion between segments of a scene with respect to the observer (camera) such as due to quick movement of a scene segment toward or away from the observer.  Therefore, weighted spatial localization of the receptive field is important for flow estimation because of the following reasons.
    
    First, the matching pixels in the subsequent frames for any pixels in the reference frame are expected to be closer and within a certain radius. While it is important for a method to look far enough in the subsequent frames to account for large motions, the matching pursuit should give priority to matching pixels that are near the reference pixels. This property is also essential for achieving high contrast motion estimates.  Second, the brightness constancy constraints typically employed in non-learning-based optical flow estimation methods are too restrictive and are likely violated in real-world sequences.
    
    Therefore, similar to patch-matching methods, an optimal matching is achieved between a reference pixel and candidate matching pixels when there is also flow agreement among neighboring pixels.  In classical optical flow methods, this is achieved through a smoothness constraint on the flow estimates.  Our earlier experiments revealed that directly enforcing this smoothness constraint in the cost function is not useful for learning-based methods and introduces difficulties in correctly estimating displacements, especially in the motion boundaries.  These two somewhat contradictory requirements suggest that a Gaussian-shaped ERF is ideally suited for optical flow estimation. Therefore, a sufficiently larger Gaussian ERF could be useful for estimating larger flows.  And by the nature of its spatial weighting of matching pixels based on their distance from a reference pixel, a Gaussian ERF could provide sharper motion boundaries in the flow estimates.
    
\subsection{Deep Dilated Convolutional Neural Network Basic: DDCNet-B0}
    Based on the desired characteristics of the learning-based models discussed earlier, we devise the following criteria to guide our network designs.

    \begin{enumerate}
        \item \emph{ERF Extent}: FWHM of ERF map should cover majority of motion vectors in the input training data.
        
        \item \emph{ERF Shape}: ERF of the untrained network should be Gaussian without significant gridding artifacts. It should be noted that fine gridding may be introduced by the network after training to achieve high contrast motion boundaries.
        
        \item \emph{Feature Resolution}: Because spatial information is critical for optical flow estimation, we propose to avoid pooling layers in the network as much as possible.  We expect that this will retain the features at their original resolution for eventual summative processing.
        
        \item \emph{Compactness}: Achieve a compact network by using relatively fewer layers and fewer filters to minimize the number of trainable parameters, accelerate training time and testing time.
        
        \item \emph{End-to-end training}: After designing the network, minimum user supervision should be used to train the network. We will not apply any post-processing on the results of the network. Entire training should be performed in an end-to-end fashion.
    \end{enumerate}
    
    Our basic learning-based optical flow estimation model receives two consecutive frames as input and generates optical flow estimates as output. As per our design criteria, we do not use any pooling layers to avoid losing spatial resolution. Therefore, all the convolutions are with stride one and there is no downsampling layer in the network. Two main parameters of our design were the number of layers and dilation factors of each of the convolution layers. Based on our findings, we increased the dilation rates linearly across the network to minimize gridding artifacts and to achieve a Gaussian ERF.  The dilation rate was set to 1 for the first layer and was increased by 1 for each of the subsequent layers.

    \begin{figure}[htbp]
        \centering
        \subcaptionbox{Histogram of flow vectors \label{fig:MagnitudeHistVsFWHM:a}}{\scalebox{0.52}{\begin{tikzpicture}
\begin{axis}[
colorbar,
colorbar style={ytick={1,2,3,4,5,6,7,8},yticklabels={\(\displaystyle {10^{1}}\),\(\displaystyle {10^{2}}\),\(\displaystyle {10^{3}}\),\(\displaystyle {10^{4}}\),\(\displaystyle {10^{5}}\),\(\displaystyle {10^{6}}\),\(\displaystyle {10^{7}}\),\(\displaystyle {10^{8}}\)},ylabel={Count}},
colormap/blackwhite,
point meta max=8.15,
point meta min=0.45,
tick align=outside,
tick pos=left,
x grid style={white!69.0196078431373!black},
xlabel={U},
xmin=-199.5, xmax=198.5,
xtick style={color=black},
y grid style={white!69.0196078431373!black},
ylabel={V},
ymin=-199.5, ymax=198.5,
ytick style={color=black}
]
\addplot graphics [includegraphics cmd=\pgfimage,xmin=-199.5, xmax=198.5, ymin=-199.5, ymax=198.5] {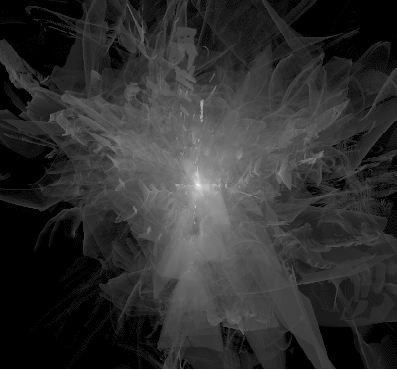};
\end{axis}
\end{tikzpicture}%
}}
        \subcaptionbox{FWHM of DDCNet-B0 \label{fig:MagnitudeHistVsFWHM:b}}{\scalebox{0.48}{\input{Figures/ERF/B0/1.tex}}}
        \decoRule
        \caption[Histogram of flow vectors in training data vs FWHM of DDCNet-B0]{Histogram of endpoint flow vectors (log scale) in Sintel training data vs Full-Width Half Maximum (FWHM) of DDCNet-Basic0 (B0). Using the histogram of flow magnitudes of training image sequences shown in Figure \subref{fig:MagnitudeHistVsFWHM:a}, the depth of the network was increased until the FWHM of the network ERF, shown in \subref{fig:MagnitudeHistVsFWHM:b}, covered the majority of the flow magnitudes in the training data.}
        \label{fig:MagnitudeHistVsFWHM}
    \end{figure}
    
\begin{figure}[htbp]
        \centering
        \scalebox{0.8}{\input{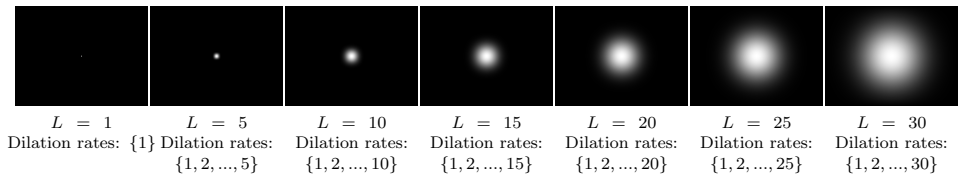}}
        \decoRule
        \caption[Effect of increasing dilation rates on ERF]{Effect of having more convolutional layers with increasing dilation rates on ERF. Each image shows the effect of the number of layers $L$ on the network ERF. Starting with a dilation rate of 1 for the first layer, the dilation rates of each of the subsequent layers were increased in steps of 1 up to the $L$th layer. For example, the network corresponding to the left-most ERF had 1 convolutional layer with a dilation rate of 1, and the network corresponding to the rightmost ERF had 30 convolutional layers with dilation rates from 1 through 30 in steps of 1.  The network (DDCNet-B0) with 25 layers was chosen as an optimal choice based on our design criteria.}
        \label{fig:IncrementalDilationRateERF}
    \end{figure}

    The optimal number of dilated convolutional layers was determined as the minimum number of dilated convolutional layers required to achieve a network whose FWHM of the ERF was large enough to cover the maximum flow magnitudes in the training data (see Figure~\ref{fig:MagnitudeHistVsFWHM}). Figure \ref{fig:IncrementalDilationRateERF} shows the receptive field for different networks as a function of the number of layers in the network. With 25 dilated convolutional layers, the spatial extent of the network ERF was sufficient to estimate larger flow velocities from the image sequences.
    
\begin{figure}[htbp]
        \centering
        \scalebox{0.7}{\def\minWidth{0.32}%
\def\minHeight{5}%
\def\layerDistance{2.2}
\def\nodeDistance{1.3}
\trimbox{0.5cm 1cm 1cm 0cm}{
\begin{tikzpicture}[
     shorten >=1pt,->,
     draw=black!50,
     node distance=\layerDistance,
     every pin edge/.style={<-,shorten <=1pt},
     ]
     \tikzstyle{neuron} = [rectangle,fill=black!25,minimum width=\minWidth cm, minimum height=\minHeight cm, inner sep=0pt, anchor=center, font=\footnotesize]
    \tikzstyle{inputNeuronStyle} = [neuron, fill=green!50]
    \tikzstyle{outputNeuronStyle}=[neuron, fill=red!50]
    \tikzstyle{hiddenNeuronStyle}=[neuron, fill=blue!50]
    \tikzstyle{zeroNeuronStyle}=[neuron, fill=gray!20]
    \tikzstyle{annotatonStyle}=[text width=3cm,text centered, scale=1, rotate=90]
    \tikzstyle{convInfoStyle}=[text width=4cm, text centered, rotate=0, scale=0.9]
    \tikzstyle{dotLayerStyle}=[text width=3cm,text centered, scale=1]
    \def\hw{\ensuremath{h, w}}
    \def\fNum{64}
    \def\id{0}
    \def\startx{0}
    \def\starty{0}
    \layerFlat{\id}{\startx-\minWidth-0.05}{\starty}{}{Frame 1}{$(\hw, 3)$}{inputNeuronStyle}
    \layerFlat{\id}{\startx}{\starty}{}{Frame 2}{$(\hw, 3)$}{inputNeuronStyle}
    \def\hiddenLayerNum{2}
    \foreach \id in {1,...,\hiddenLayerNum}{
        \pgfmathparse{\layerDistance*\id-0.5*\layerDistance}
        \edef\startx{\pgfmathresult}
        \pgfmathparse{-0.5*\minHeight}
        \edef\tmpy{\pgfmathresult}
        \textOnlyLayer{\id}{\startx}{\tmpy}{Conv \id \\ \filterNum{\id}=\fNum \\ \filterDilation{\id}=[\id,\id]\\ \activationFunction{\id}=ReLU}
        \pgfmathparse{\layerDistance*\id}
        \edef\startx{\pgfmathresult}
        \layerFlat{\id}{\startx}{\starty}{}{Feature map \id}{$(\hw, \fNum)$}{hiddenNeuronStyle}
    }
    \pgfmathparse{\layerDistance*\hiddenLayerNum + 0.5*\layerDistance}
    \edef\startx{\pgfmathresult}
    \pgfmathparse{-0.5*\minHeight}
    \edef\starty{\pgfmathresult}
    \pgfmathparse{-\minHeight - \nodeDistance}
    \edef\tempy{\pgfmathresult}
    \dotlayer{\id-dot}{\startx}{\starty}{\tempy}{dotLayerStyle}
    \def\starty{0}
    \def\id{24}
    \pgfmathparse{\startx+0.5*\layerDistance}
    \xdef\startx{\pgfmathresult}
    \layerFlat{\id}{\startx}{\starty}{}{Feature map \id}{$(\hw, \fNum)$}{hiddenNeuronStyle}
    \foreach \id in {25}{
            \pgfmathparse{\startx+0.5*\layerDistance}
            \xdef\startx{\pgfmathresult}
            \pgfmathparse{-0.5*\minHeight}
            \edef\tmpy{\pgfmathresult}
            \textOnlyLayer{\id}{\startx}{\tmpy}{Conv \id \\ \filterNum{\id}=\fNum  \\ \filterDilation{\id}=[\id,\id] \\ \activationFunction{\id}=ReLU}
            \pgfmathparse{\startx+0.5*\layerDistance}
            \xdef\startx{\pgfmathresult}
            \layerFlat{\id}{\startx}{\starty}{}{Feature map \id}{$(\hw, \fNum)$}{hiddenNeuronStyle}
        }
    \foreach \id/\dilation in {26/12, 27/6, 28/3, 29/1}{
            \pgfmathparse{\startx+0.5*\layerDistance}
            \xdef\startx{\pgfmathresult}
            \pgfmathparse{-0.5*\minHeight}
            \edef\tmpy{\pgfmathresult}
            \textOnlyLayer{\id}{\startx}{\tmpy}{Conv \id \\ \filterNum{\id}=\fNum \\ \filterDilation{\id}=[\dilation,\dilation] \\ \activationFunction{\id}=ReLU}
            \pgfmathparse{\startx+0.5*\layerDistance}
            \xdef\startx{\pgfmathresult}
            \layerFlat{\id}{\startx}{\starty}{}{Feature map \id}{$(\hw, \fNum)$}{hiddenNeuronStyle}
        }
    \def\fNum{2}
    \def\id{30}
    \pgfmathparse{\startx+0.5*\layerDistance}
    \xdef\startx{\pgfmathresult}
    \pgfmathparse{-0.5*\minHeight}
    \edef\tmpy{\pgfmathresult}
    \textOnlyLayer{\id}{\startx}{\tmpy}{Conv \id \\ \filterNum{\id}=\fNum \\ \filterDilation{\id}=[1,1] \\ \activationFunction{\id}=Linear}
    \pgfmathparse{\startx+0.5*\layerDistance}
    \xdef\startx{\pgfmathresult}
    \layerFlat{\id}{\startx}{\starty}{}{UV map}{$(\hw, \fNum)$}{outputNeuronStyle}
    \foreach \i in {0,...,1}{
        \pgfmathparse{int(\i + 1)}
        \edef\j{\pgfmathresult}
        \draw [line width=1,->, shorten >=0.5cm, shorten <=0.5cm] (\i-nb) edge (\j-nb);
        }
    \foreach \i in {24,...,29}{
        \pgfmathparse{int(\i + 1)}
        \edef\j{\pgfmathresult}
        \draw [line width=1,->, shorten >=0.5cm, shorten <=0.5cm] (\i-nb) edge (\j-nb);
    }
\end{tikzpicture}
}}
        \decoRule
        \caption[Basic \methodname]{Architecture of our basis deep dilated CNN (DDCNet-B0) for optical flow estimation.$\filterNum{\ell}$ is the number of filters, $\filterDilation{\ell}$ is the dilation rate, and $\activationFunction{\ell}$ is the activation function of layer $\ell$.}
        \label{fig:DDCNetB0}
    \end{figure}
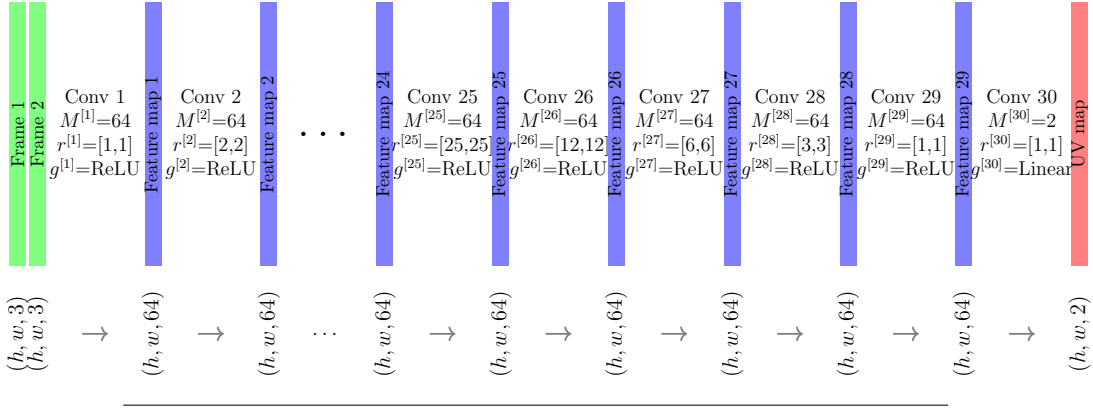

Figure~\ref{fig:DDCNetB0} shows the architecture of our basic deep dilated CNN (DDCNet-B0) for optical flow estimation. It has 25 layers of dilated convolutions with dilation rates from 1 through 25 in steps of 1. To improve the accuracy of flow estimates and motion boundaries, we appended the network with four more layers with decreasing dilation rates of $(12, 6, 3, 1)$. Except for the last layer, there are 64 $3\times3$ filters in each of these layers with a \textit{ReLU} activation function. The last layer is a convolution layer (dilation rate is 1) with two $1\times1$ filters and a linear activation function to generate the final flow estimates (UV-map). In this and other following network diagrams,  $\filterNum{\ell}$ is the number of filters, $\filterDilation{\ell}$ is the dilation rate, and $\activationFunction{\ell}$ denotes activation function in layer $\ell$. 

\textit{Endpoint Error} characterizes network performance at a given pixel location $(i, j)$ within an image sequence and is estimated as the Euclidean distance between the ground truth $\Vec{f}_g$ and estimated flow vector $\Vec{f}_e$ at that location as follows.

    \begin{equation}\label{eq:endpointerror}
    \begin{split}
        EE(i,j) & = \lVert \Vec{f}_g(i,j) - \Vec{f}_e(i,j) \rVert_2 \\  
                & =  \sqrt{(u_e(i,j) - u_g(i,j))^2 + (v_e(i,j) - v_g(i,j))^2}
    \end{split}
    \end{equation}
    
	\textit{Average Endpoint Error} (AEE) characterizes network performance at all pixel locations within an image sequence.  We use AEE metric for training the networks and is estimated as the average of EEs at all pixel locations as follows.
	
	\begin{equation}\label{eq:averageendpointerror}
	    	AEE = \frac{1}{h \times w} \sum_i^{h} \sum_{j}^{w} EE(i,j)
	\end{equation}
	
	An average AEE measure based on the AEEs of all image sequences in a benchmark dataset is used for assessing the network performance on benchmark datasets.
    
\subsection{Strategies for Improving DDCNet-B0 Network}
    To increase the accuracy of the method and reduce the training time, we considered further improvements to our basic network described in the previous section.  Adding more layers, more filters, and stacking several sub-networks on top of each other are some of the common strategies available for improving the accuracy of learning-based methods \cite{ilg2017flownet}.  These strategies, however, require substantially more memory and computational resources. Therefore, we retain our focus on \emph{simplicity} and \emph{compactness} as we systematically improve our basic network architecture DDCNet-B0. 
    
    \subsection{Spatial Feature Extractor Module}
    Raw pixel intensities of the image sequences are not ideal for matching pursuits as they are strongly influenced by changes in scene illumination and scene orientation. As can be observed from our preliminary results of DDCNet-B0, the network was able to internally extract relevant and robust features, identify pixel correspondences between frames, and estimate the displacement vectors using only the \textit{raw input images} and a suitable cost function for optimization and learning. 
    
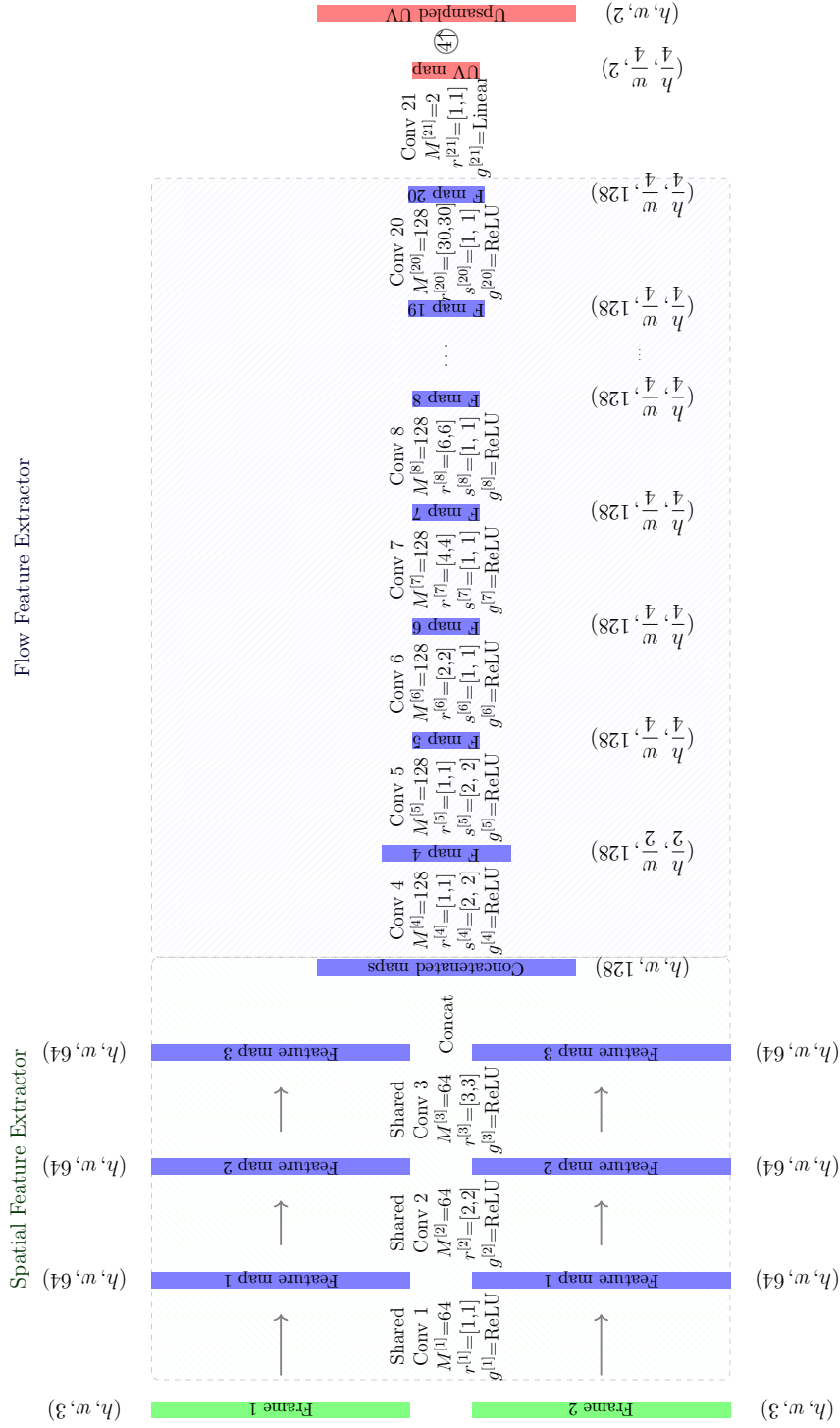
\begin{sidewaysfigure}
        \centering
        \scalebox{0.7}{\def\minWidth{0.32}%
\def\minHeight{5}%
\def\layerDistance{2.2}
\def\nodeDistance{1.3}
           
\trimbox{0.5cm 1cm 1cm 0cm}{
\begin{tikzpicture}[
     shorten >=1pt,->,
     draw=black!50,
     node distance=\layerDistance,
     every pin edge/.style={<-,shorten <=1pt},
     ]
     \tikzstyle{neuron} = [rectangle,fill=black!25,minimum width=\minWidth cm, minimum height=\minHeight cm, inner sep=0pt, anchor=center, font=\footnotesize]
    \tikzstyle{inputNeuronStyle} = [neuron, fill=green!50]
    \tikzstyle{outputNeuronStyle}=[neuron, fill=red!50]
    \tikzstyle{hiddenNeuronStyle}=[neuron, fill=blue!50]
    \tikzstyle{zeroNeuronStyle}=[neuron, fill=gray!20]
    \tikzstyle{annotatonStyle}=[text width=3cm,text centered, scale=1, rotate=90]
    \tikzstyle{convInfoStyle}=[text width=4cm, text centered, rotate=0, scale=0.9]
    \tikzstyle{dotLayerStyle}=[text width=3cm,text centered, scale=0.5]
    \def\hw{\ensuremath{h, w}}
    \def\fNum{64}
    \def\startx{-\minWidth}
    \def\starty{0}
    \layerFlat{f-1}{\startx}{\starty}{$(\hw, 3)$}{Frame 1}{}{inputNeuronStyle}
    
    \def\preprocessLayerNum{3}
    \foreach \id in {1,...,\preprocessLayerNum}{
        \pgfmathparse{\layerDistance*\id}
        \edef\tmpx{\pgfmathresult}
        \layerFlat{p-1-\id}{\tmpx}{\starty}{$(\hw, \fNum)$}{Feature map \id}{}{hiddenNeuronStyle}
    }
    
    \def\starty{-1.5*\minHeight}
    \layerFlat{f-2}{\startx}{\starty}{}{Frame 2}{$(\hw, 3)$}{inputNeuronStyle}
    
    \foreach \id in {1,...,\preprocessLayerNum}{
        \pgfmathparse{\layerDistance*\id}
        \edef\tmpx{\pgfmathresult}
        \layerFlat{p-2-\id}{\tmpx}{\starty}{}{Feature map \id}{$(\hw, \fNum)$}{hiddenNeuronStyle}
    }
    
    \def\starty{-1.5*\minHeight}
    \foreach \id in {1,...,\preprocessLayerNum}{
        \pgfmathparse{\layerDistance*\id-0.5*\layerDistance}
        \xdef\startx{\pgfmathresult}
        \textOnlyLayer{\id}{\startx}{\starty+0.5}{Shared \\ Conv \id \\ \filterNum{\id}=\fNum \\ \filterDilation{\id}=[\id,\id]\\ \activationFunction{\id}=ReLU}
    }
    
     \def\starty{-0.9*\minHeight}
    \pgfmathparse{\startx+0.75*\layerDistance}
    \xdef\startx{\pgfmathresult}
    \textOnlyLayer{c}{\startx}{\starty-0.5*\minHeight}{Concat}
    \pgfmathparse{\startx+0.5*\layerDistance}
    \xdef\startx{\pgfmathresult}
    \pgfmathparse{int(2*\fNum)}
    \edef\chl{\pgfmathresult}
    \layerFlat{conc}{\startx}{\starty}{}{Concatenated maps}{$(\hw, \chl)$}{hiddenNeuronStyle}
    
    \def\fNum{128}
    \def\p{0.25}
    \def\st{[1, 1]}
    \def\hw{{\dfrac{h}{4}, \dfrac{w}{4}}}
    \foreach \id/\dilation in {4/1, 5/1, 6/2, 7/4, 8/6}{
            \ifnum \id=4
                \edef\p{0.5}
                \def\st{[2, 2]}
                    \def\hw{{\dfrac{h}{2}, \dfrac{w}{2}}}
            \fi
            \ifnum  \id=5
                \def\st{[2, 2]}
            \fi
            \pgfmathparse{\startx+0.5*\layerDistance}
            \xdef\startx{\pgfmathresult}
            \textOnlyLayer{\id}{\startx}{\starty-0.5*\minHeight}{Conv \id \\ \filterNum{\id}=\fNum \\ \filterDilation{\id}=[\dilation,\dilation]\\
            \filterStride{\id}=\st \\\activationFunction{\id}=ReLU}
            \tikzstyle{hiddenNeuronStyle}=[neuron, fill=blue!50,  minimum height=\p*\minHeight cm]
            \pgfmathparse{\startx+0.5*\layerDistance}
            \xdef\startx{\pgfmathresult}
            \layerFlat{\id}{\startx}{\starty}{}{F map \id}{$(\hw, \fNum)$}{hiddenNeuronStyle}
        }
    
    \pgfmathparse{\startx + 0.4*\layerDistance}
    \edef\startx{\pgfmathresult}
    \dotlayer{dot}{\startx}{\starty - 0.5*\minHeight}{\starty- 1.25*\minHeight}{dotLayerStyle}
    
    \pgfmathparse{\startx - 0.1*\layerDistance}
    \edef\startx{\pgfmathresult}
    \foreach \id/\dilation in {19/28, 20/30}{
            \ifnum \id=20
                \pgfmathparse{\startx+0.5*\layerDistance}
                \xdef\startx{\pgfmathresult}
                \textOnlyLayer{\id}{\startx}{\starty-0.5*\minHeight}{Conv \id \\ \filterNum{\id}=\fNum \\ \filterDilation{\id}=[\dilation,\dilation]\\
            \filterStride{\id}=\st \\\activationFunction{\id}=ReLU}
            \fi
            \tikzstyle{hiddenNeuronStyle}=[neuron, fill=blue!50,  minimum height=\p*\minHeight cm]
            \pgfmathparse{\startx+0.5*\layerDistance}
            \xdef\startx{\pgfmathresult}
            \layerFlat{\id}{\startx}{\starty}{}{F map \id}{$(\hw, \fNum)$}{hiddenNeuronStyle}
        }

    \def\fNum{2}
    \def\id{21}
    \pgfmathparse{\startx+0.6*\layerDistance}
    \xdef\startx{\pgfmathresult}
    \textOnlyLayer{\id}{\startx}{\starty - 0.5*\minHeight}{Conv \id \\ \filterNum{\id}=\fNum \\ \filterDilation{\id}=[1,1] \\ \activationFunction{\id}=Linear }
    \pgfmathparse{\startx+0.5*\layerDistance}
    \xdef\startx{\pgfmathresult}
    \tikzstyle{outputNeuronStyle}=[neuron, fill=red!50,  minimum height=\p*\minHeight cm]
    \layerFlat{\id}{\startx}{\starty}{}{UV map}{$(\hw, \fNum)$}{outputNeuronStyle}

    \def\hw{\ensuremath{h, w}}
    \pgfmathparse{\startx+0.25*\layerDistance}
    \xdef\startx{\pgfmathresult}
    
    
    \draw[thick] (\startx,\starty - 0.5*\minHeight) circle [radius=0.23] node {${4}{\uparrow}$};
    
    \pgfmathparse{\startx+0.25*\layerDistance}
    \xdef\startx{\pgfmathresult}
    \tikzstyle{outputNeuronStyle}=[neuron, fill=red!50,  minimum height=\minHeight cm]
    \layerFlat{up}{\startx}{\starty}{}{Upsampled UV}{$(\hw, \fNum)$}{outputNeuronStyle}
    
    \draw [line width=1,->, shorten >=0.5cm, shorten <=0.5cm] (f-1) edge (p-1-1);
    \draw [line width=1,->, shorten >=0.5cm, shorten <=0.5cm] (f-2) edge (p-2-1);
     \foreach \i in {1, 2}{
         \pgfmathparse{int(\i + 1)}
         \edef\j{\pgfmathresult}
         \draw [line width=1,->, shorten >=0.5cm, shorten <=0.5cm] (p-1-\i) edge (p-1-\j);
          \draw [line width=1,->, shorten >=0.5cm, shorten <=0.5cm] (p-2-\i) edge (p-2-\j);
      }
    \begin{pgfonlayer}{background}
    \edef\x{2*\layerDistance}
    \edef\y{-9.7*\nodeDistance}
    \edef\w{3.7*\layerDistance}
    \edef\h{8.8*\nodeDistance cm}
        
    \coordinate (A) at (0.2*\nodeDistance, -\nodeDistance);
    \coordinate (B) at (3.84*\layerDistance, -9.61*\nodeDistance);
    \coordinate (C) at (10.7*\layerDistance, -\nodeDistance);
    \node[moduleBox={(A) (B)}, label={[label distance=1.7*\nodeDistance cm, color=green!20!black]90:Spatial Feature Extractor}, pattern=north west lines, pattern color=green] {}; 
    \node[moduleBox={(B) (C)}, label={[label distance=1.7*\nodeDistance cm, color=blue!20!black]90:Flow Feature Extractor}, pattern=north east lines, pattern color=blue] {}; 
    \end{pgfonlayer}
    
\end{tikzpicture}
}}
        \decoRule
        \caption[Initial design of DDCNet-Basic1]{Improvements to DDCNet-B0: A \textit{spatial feature extractor} module (highlighted by light green) is added to the initial design DDCNet-B0 to guide the network to extract illumination invariant image features before learning to do the matching task. The second part of the design improvement, highlighted in light blue color, is called the \textit{flow feature extractor module}. Compared to DDCNet-B0, the resolution of feature maps is reduced by a factor of four to increase ERF and the dilation rates were increased in steps of 2 to reduce the processing overhead.}
        \label{fig:DDCNetBasic1Initial}
    \end{sidewaysfigure}

    One possible avenue to improve the accuracy of the estimates is by directly allowing the network to extract and utilize illumination-independent image sequences for flow estimation. The network estimation accuracy slightly improved when gradient image sequences (spatial derivatives) along with raw image sequences were used as input to the same network, thus validating the merits of this approach. The gradient images likely provided more robust features less dependent on temporal noise and illumination changes for flow estimation. By decoupling spatial feature extraction steps from temporal matching steps in our basic network design DDCNet-B0, the network can possibly learn these spatial derivatives and other space-invariant  features before matching pursuits. Therefore, we added a 3-layer section in the network for this preprocessing task which takes each image separately and generates 64 feature maps for further processing. The preprocessing section is shared between the first and second frames of the image sequence \ie same filters are used for both images in the input image sequence. We refer to this preprocessing part of the network as a \emph{spatial feature extractor} and the corresponding network segment is highlighted in light green color in Figure~\ref{fig:DDCNetBasic1Initial}.
    
\subsection{Flow Feature Extractor Module}
    A second major design improvement to DDCNet-B0 is to reduce the computational demands by decreasing the depth of the network while maintaining its receptive field and accuracy. This is achieved by dropping every other layer from the DDCNet-B0. Our experiments show that an even step size of dilation rates yields better results compared to odd step sizes. By adjusting the number of layers to have the desired ERF, we ended up with a 15-layer sub-net with the dilation rates of ${1, 2, 4, 6, \ldots, 30}$. The resulting network segment that we refer to as a \emph{flow feature extractor} module is highlighted in blue color in Figure~\ref{fig:DDCNetBasic1Initial}.
    
    To further reduce the processing time and increase the ERF of the network, we reduce the spatial dimensions of feature maps in the first and second layers of the \emph{flow feature extractor} part by using a stride length of 2 for convolutions with an effective downsampling rate of 4:1. These parameters were experimentally determined to provide a balance between reduction in computational load and preserving spatial resolution of the feature maps.
    
    \subsection{Improved DDCNet Basic: DDCNet-B1}
\begin{sidewaysfigure}[htbp]
        \centering
        \scalebox{0.8}{\def\minWidth{0.32}%
\def\minHeight{5}%
\def\layerDistance{2.3}
\def\nodeDistance{1.3}
\trimbox{0.5cm 1cm 1cm 0cm}{
\begin{tikzpicture}[
     shorten >=1pt,->,
     draw=black!50,
     node distance=\layerDistance,
     every pin edge/.style={<-,shorten <=1pt}
     ]
     \tikzstyle{neuron} = [rectangle,fill=black!25,minimum width=\minWidth cm, minimum height=\minHeight cm, inner sep=0pt, anchor=center, font=\footnotesize]
    \tikzstyle{inputNeuronStyle} = [neuron, fill=green!50]
    \tikzstyle{outputNeuronStyle}=[neuron, fill=red!50]
    \tikzstyle{hiddenNeuronStyle}=[neuron, fill=blue!50]
    \tikzstyle{hiddenNeuronStyleFourth}=[hiddenNeuronStyle, minimum height=0.25*\minHeight cm]
    \tikzstyle{zeroNeuronStyle}=[neuron, fill=gray!20]
    \tikzstyle{annotatonStyle}=[text width=3cm,text centered, scale=1, rotate=90]
    \tikzstyle{convInfoStyle}=[text width=4cm, text centered, rotate=0, scale=0.9]
    \tikzstyle{dotLayerStyle}=[text width=3cm,text centered, scale=1]
    \tikzstyle{moduleBlock} = [rectangle, draw, inner sep=0pt, dashed, rounded corners,opacity=0.3,fit=#1]
    
    \def\hw{\ensuremath{h, w}}
    \def\fNum{64}
    \def\startx{-\minWidth}
    \def\starty{0}
    \layerFlat{f-1}{\startx}{\starty}{$(\hw, 3)$}{Frame 1}{}{inputNeuronStyle}
    
    \def\starty{-1.5*\minHeight}
    \layerFlat{f-2}{\startx}{\starty}{}{Frame 2}{$(\hw, 3)$}{inputNeuronStyle}
    
    \pgfmathparse{\startx+0.25*\layerDistance}
    \xdef\startx{\pgfmathresult}
    
    \def \blockTopY{-\nodeDistance}
    \def \blockBottomY{-9.61*\nodeDistance}
    \def \blockStartX{\startx}
    \def \blockSize{0.75*\layerDistance}
    \coordinate (A) at (\blockStartX, \blockTopY);
    \coordinate (B) at (\blockStartX+\blockSize, \blockBottomY);
    \node[name=block0, moduleBlock={(A) (B)}, label={[rotate=90, xshift=0*\layerDistance cm, color=black]center:Spatial Feature Extractor}, pattern= crosshatch dots, pattern color=green] {};
    \draw [line width=1,->] (f-1) -- +(0:1.5*\minWidth);
    \draw [line width=1,->] (f-2) -- +(0:1.5*\minWidth);
    
     \def\starty{-0.9*\minHeight}
     \pgfmathparse{\startx+ \blockSize + 0.58*\minWidth}
     \xdef\startx{\pgfmathresult}
     \pgfmathparse{int(2*\fNum)}
     \edef\chl{\pgfmathresult}
     \layerFlat{map0}{\startx}{\starty}{}{Spatial feature maps}{$(\hw, \chl)$}{hiddenNeuronStyle}
     
    \def \blockStartX{\startx+0.25*\layerDistance}
    \def \blockSize{1.25*\layerDistance}
    \coordinate (A) at (\blockStartX, \blockTopY);
    \coordinate (B) at (\blockStartX+\blockSize, \blockBottomY);
    \node[name=block1, moduleBlock={(A) (B)}, label={[rotate=90, xshift=0*\layerDistance cm, color=black]center:Flow Feature Extractor}, pattern=crosshatch dots, pattern color=blue] {}; 
     \draw [line width=1,->] (map0) -- +(0:1*\minWidth);   
     
      \def\p{0.25}
      \def\hw{{\frac{h}{4}, \frac{w}{4}}}
      \def\starty{-0.9*\minHeight}
      \pgfmathparse{\startx+ \blockSize + 2.45*\minWidth}
      \xdef\startx{\pgfmathresult}
      \pgfmathparse{int(2*\fNum)}
      \edef\chl{\pgfmathresult}
      \layerFlat{map18}{\startx}{\starty}{}{F map 18}{$(\hw, \chl)$}{hiddenNeuronStyleFourth}
      
      \draw [line width=1,->] (map18) -- +(0:1*\minWidth);  
        
    \def\starty{-0.9*\minHeight}

    \def\p{0.5}
    \def\hw{{\frac{h}{2}, \frac{w}{2}}}
    \pgfmathparse{\startx+0.27*\layerDistance}
    \xdef\startx{\pgfmathresult}
    \draw[thick] (\startx,\starty - 0.5*\minHeight) circle [radius=0.23] node {${2}{\uparrow}$};
    
    \pgfmathparse{\startx+0.25*\layerDistance}
    \xdef\startx{\pgfmathresult}
    \tikzstyle{hiddenNeuronStyle}=[neuron, fill=blue!50,  minimum height=\p*\minHeight cm]
    \layerFlat{conc}{\startx}{\starty}{}{Upsampled maps}{$(\hw, \fNum)$}{hiddenNeuronStyle}
    
    \def\fNum{64}
    \def\st{[1, 1]}
    \def\hw{{\frac{h}{2}, \frac{w}{2}}}
    \foreach \id/\dilation in {19/1, 20/2, 21/3}{
            \pgfmathparse{\startx+0.5*\layerDistance}
            \xdef\startx{\pgfmathresult}
            \textOnlyLayer{\id}{\startx}{\starty-0.5*\minHeight}{Conv \id \\ \filterNum{\id}=\fNum \\ \filterDilation{\id}=[\dilation,\dilation]\\
            \filterStride{\id}=\st \\\activationFunction{\id}=ReLU}
            \tikzstyle{hiddenNeuronStyle}=[neuron, fill=blue!50,  minimum height=\p*\minHeight cm]
            \pgfmathparse{\startx+0.5*\layerDistance}
            \xdef\startx{\pgfmathresult}
            \layerFlat{\id}{\startx}{\starty}{}{F map \id}{$(\hw, \fNum)$}{hiddenNeuronStyle}
        }
    
    \pgfmathparse{\startx + 0.5*\layerDistance}
    \edef\startx{\pgfmathresult}
    \dotlayer{dot}{\startx}{\starty - 0.5*\minHeight}{\starty- 1.25*\minHeight}{dotLayerStyle}
    
    \foreach \id/\dilation in {27/9, 28/10}{
            \ifnum \id=28
                \pgfmathparse{\startx+0.5*\layerDistance}
                \xdef\startx{\pgfmathresult}
                \textOnlyLayer{\id}{\startx}{\starty-0.5*\minHeight}{Conv \id \\ \filterNum{\id}=\fNum \\ \filterDilation{\id}=[\dilation,\dilation]\\
            \filterStride{\id}=\st \\\activationFunction{\id}=ReLU}
            \fi
            \tikzstyle{hiddenNeuronStyle}=[neuron, fill=blue!50,  minimum height=\p*\minHeight cm]
            \pgfmathparse{\startx+0.5*\layerDistance}
            \xdef\startx{\pgfmathresult}
            \layerFlat{\id}{\startx}{\starty}{}{F map \id}{$(\hw, \fNum)$}{hiddenNeuronStyle}
        }

    \def\fNum{2}
    \def\id{29}
    \pgfmathparse{\startx+0.55*\layerDistance}
    \xdef\startx{\pgfmathresult}
    \textOnlyLayer{\id}{\startx}{\starty - 0.5*\minHeight}{Conv \id \\ \filterNum{\id}=\fNum  \\ \filterDilation{\id}=[1,1] \\ \activationFunction{\id}=Linear}
    \pgfmathparse{\startx+0.5*\layerDistance}
    \xdef\startx{\pgfmathresult}
    \tikzstyle{outputNeuronStyle}=[neuron, fill=red!50,  minimum height=\p*\minHeight cm]
    \layerFlat{\id}{\startx}{\starty}{}{UV map}{$(\hw, \fNum)$}{outputNeuronStyle}
    \draw [line width=1,->] (28) -- +(0:1*\minWidth);    
    \def\hw{\ensuremath{h, w}}
    \pgfmathparse{\startx+0.25*\layerDistance}
    \xdef\startx{\pgfmathresult}
    
    \draw[thick] (\startx,\starty - 0.5*\minHeight) circle [radius=0.23] node {${2}{\uparrow}$};
    
    \pgfmathparse{\startx+0.25*\layerDistance}
    \xdef\startx{\pgfmathresult}
    \tikzstyle{outputNeuronStyle}=[neuron, fill=red!50,  minimum height=\minHeight cm]
    \layerFlat{up}{\startx}{\starty}{}{Upsampled UV}{$(\hw, \fNum)$}{outputNeuronStyle}

    \begin{pgfonlayer}{background}
        \def \blockStartX{2.7*\layerDistance}
        \def \blockSize{5.28*\layerDistance}
        \coordinate (A) at (\blockStartX, \blockTopY);
        \coordinate (B) at (\blockStartX+\blockSize, \blockBottomY);
        \node[moduleBox={(A) (B)}, label={[label distance=0.7*\nodeDistance cm, color=blue!20!black]90:Flow Feature Refiner}, pattern=north east lines, pattern color=blue] {};
        
         \pgfmathparse{\blockStartX+\blockSize+1.8*\minWidth}
         \xdef\blockStartX{\pgfmathresult}
         \def \blockSize{1.8*\layerDistance}
         \coordinate (A) at (\blockStartX, \blockTopY);
         \coordinate (B) at (\blockStartX+\blockSize, \blockBottomY);
         \node[moduleBox={(A) (B)}, label={[label distance=0.7*\nodeDistance cm, color=blue!20!black]90:Final Estimator}, pattern=north east lines, pattern color=red] {};  
    \end{pgfonlayer}
\end{tikzpicture}
}}
        \decoRule
        \caption[DDCNet Basic1]{The full architecture of the DDCNet-B1 network. This network has an additional module called Feature Refined which consists of 10 dilated convolutional layers with increasing dilation rates from 1 to 10. This module helps the network to generate more refined estimation. $\filterNum{\ell}$ is number of filters, $\filterDilation{\ell}$ is dilation rates, and $\activationFunction{\ell}$ is activation function of layer $\ell$.}
        \label{fig:DDCNetBasic1}
    \end{sidewaysfigure}
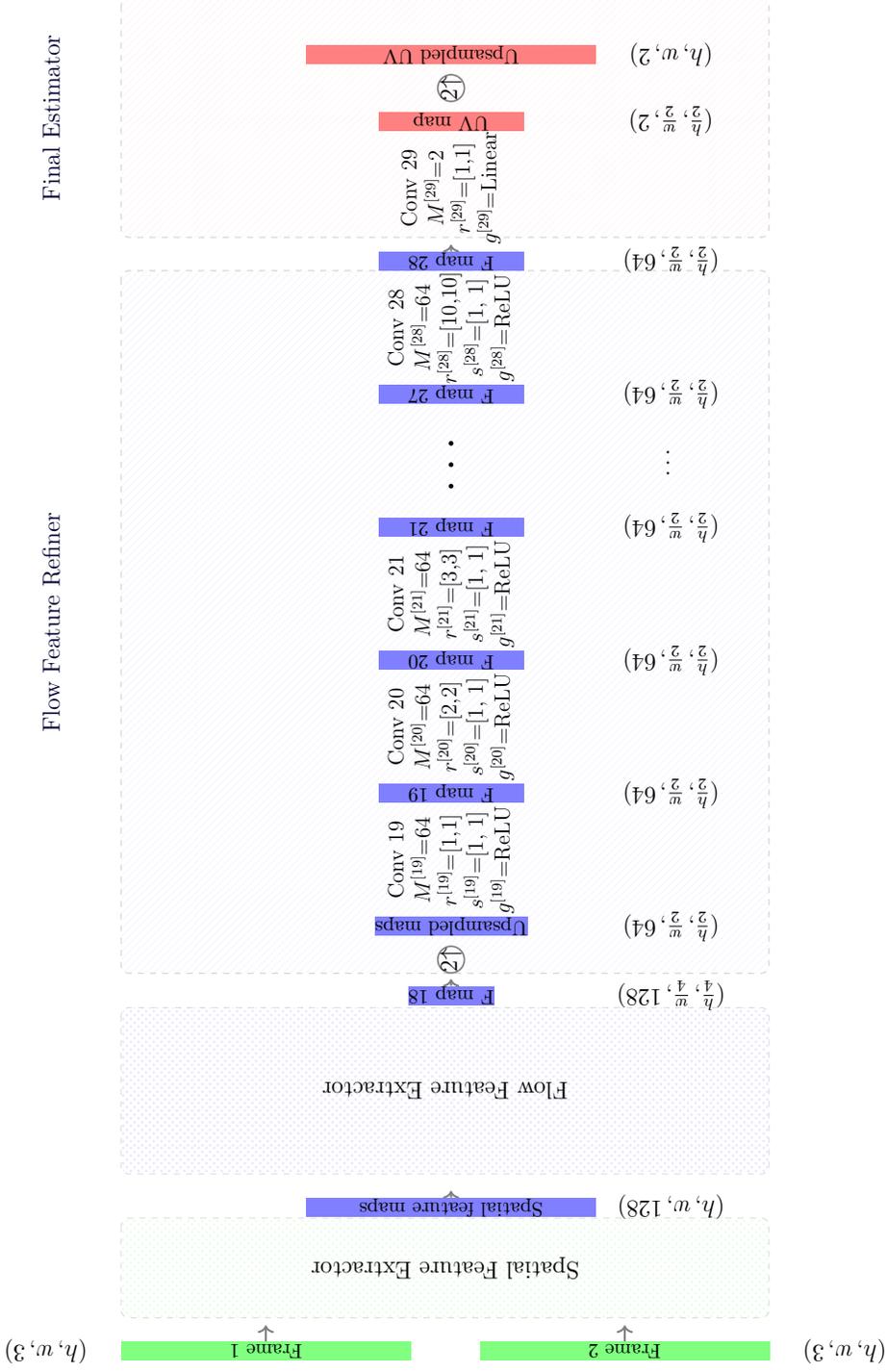

    Figure~\ref{fig:DDCNetBasic1} shows full design of the improved DDCNet-B1 network. It is worth noting that we increased the number of filters in \emph{flow feature extractor} module from 64 to 128 per layer. 
    
    With reduced-resolution feature maps, though the networks can be trained much faster, some flow details would be lost. One possibility is to simply upsample the generated flows at 1/4th resolution to the original resolution as shown in Figure~\ref{fig:DDCNetBasic1Initial}. However, it is preferred to allow the network to learn this upsampling process through training. To achieve this, we upsample the feature maps from the last layer of the \emph{flow feature extractor} module by a factor of 2 and pass it to a new \emph{feature refiner} section with increasing dilation rates of ${1, 2, 3, ..., 10}$. In other words, the features are refined and aggregated at half of the original resolution in the \emph{feature refiner} module.  Our experimental results indicated that increasing dilation rates is beneficial for this module in terms of improving the endpoint error. In contrast, decreasing dilation rates were used in the latest layers of DDCNet-B0. Using the refined features at one-half resolution, flow estimates were generated at one-half resolution and upsampled using a nearest-neighbor approach to generate the final full resolution flow estimate in the \emph{final estimator} module.

\section{Experiments}
\label{sec:experiments}
    \subsection{Network and Training Details}
    Details of each of the networks with \(3\times3\) filters in all layers are presented in Figures~\ref{fig:DDCNetB0} and \ref{fig:DDCNetBasic1}.
    
    Similar to the other methods, we utilize a simpler training dataset (in terms of types of motion as well as the texture of images in the input sequence) such as the Flying Chair sequences initially to guide the network toward an optimal network configuration for optical flow estimation. To reduce computational demand, avoid over-fitting, and enable faster convergence during training, we emphasize training of selected network segments (sub-nets) by freezing previous network sections. A batch size of 8 (if the model can be fit in the GPU memory), 4, and 2 were used for training using the $L_2$ regularization method, \emph{He} initialization method and the Adam optimizer.
    
    \textit{Train-time-augmentation} methods were used to increase the number and quality of the training dataset. Specifically, geometric augmentation using affine transformations (\eg rotation and resizing) and photometric augmentation (\eg contrast, saturation, hue, and brightness) were used. From an initial rate of $10^{-3}$, we experimentally determined a learning rate that is optimal for network convergence. The learning rate was adjusted when the rate of decrease in the training error (loss function) was smaller without further significant improvement in the network accuracy.

    \subsection{Datasets}
    A summary of various benchmark image sequences that we utilized for assessing the performance of our receptive field guided networks is presented in Table~\ref{tab:Datasets}. For each benchmark sequence, the ground truth optical flow was available.
	\begin{table}[h]
		\centering
		\caption[Summary of benchmark image sequence datasets]{Summary of benchmark image sequences used for assessing our receptive field guided network designs. \emph{GT}: Ground truth.}
		\scalebox{0.8}{\begin{tabular}{lccccc}
    \toprule
    & \tabhead{Frame}  & \tabhead{Frames}  & \tabhead{GT density} & \tabhead{Size} & \tabhead{Type}\\      
    & \tabhead{pairs}  & \tabhead{with GT} & \tabhead{per frame} & \tabhead{(\emph{Width $\times$ Height})} \\
    \midrule
    \tabhead{Middlebury}    & 72     & 8            & 100       &$388 \times 584$  & Real\\             
    \tabhead{KITTI12}       & 389    & 194          & $\sim$ 50 & $375 \times 1242$  & Real\\   
    \tabhead{KITTI15}       & 400    & 194          & $\sim$ 50 &$375 \times 1242$  & Real\\   
    \tabhead{Sintel} 	    & 2,145  & 1,041        & 100       &$436 \times 1024$  & Synthetic\\             
    \tabhead{Flying Chairs} & 22,872 & 22,872     & 100 &$384 \times 512$  & Synthetic\\ 
    \tabhead{Flying Things3D} & 77,757 & 77,757       & 100  &$540 \times 960$  & Synthetic\\ 
    \tabhead{\hspace{0.1in}3D subset} &&&&&\\
    \bottomrule
\end{tabular}}
		\label{tab:Datasets}
	\end{table}

\subsection{Results}
    Average endpoint error (AEE) in Equation~\ref{eq:averageendpointerror} was the main metric used for assessing network performance during the design and development of our networks. We also used $Fl_{\text{all}}$ on KITTI datasets, where $Fl_{\text{all}}$ measures the percentage of outlier estimates based on the number of pixels with endpoint error (EE) $\ge 3$ pixels and EE $\ge 5\%$ of the magnitude of its ground-truth flow vector were counted as outliers).
    
    \subsubsection{ERF Analysis}
\begin{figure}[htbp]
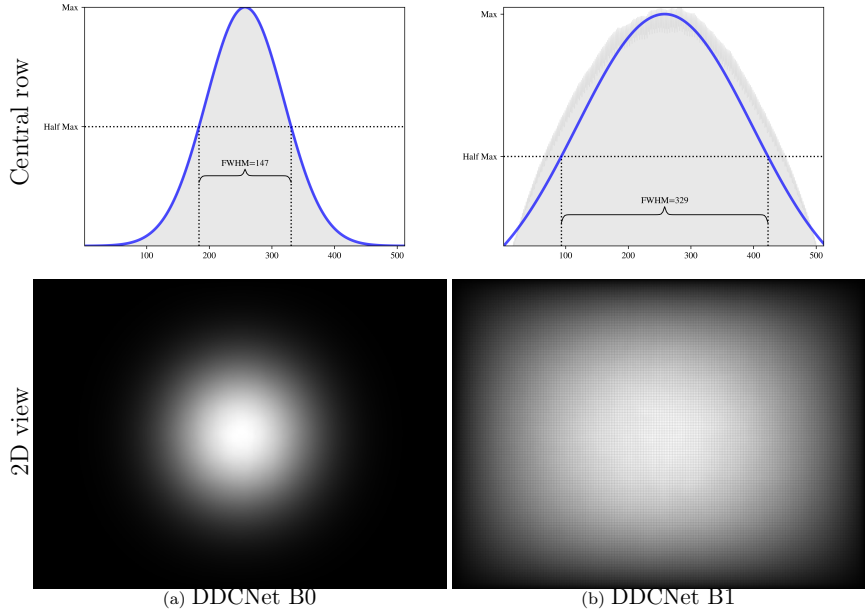

        \centering
        \scalebox{0.8}{    
    
\def\scale{0.5}
\def\rootDir{Figures/ERF}
\newcommand{\rott}[1]{\selectfont{\rotatebox[origin=c]{90}{#1}}}
\newcommand{\IMG}[1]{\includegraphics[align=c,width=\scale\textwidth]{\rootDir/#1}}

\newcommand{\rowImg}[2]{
\xdef\id{#2}
\begin{minipage}[b]{0.02\linewidth}\centering \rott{#1} \end{minipage} & \IMG{B0/\id.png} & \IMG{B1/\id.png}
}

\newcommand{\capNew}[1]{\begin{minipage}[b]{\scale\linewidth}\centering\subcaption{\small #1}\end{minipage}}
\setlength\tabcolsep{1.5pt}
\begin{tabular}{ccc}
    \rowImg{Central row}{1}\\
    \rowImg{2D view}{2}\\
    & \capNew{DDCNet B0} & \capNew{DDCNet B1}
\end{tabular}

}
        \decoRule
       \caption[ERFs of our receptive field guided networks]{ERFs of our receiptive field guided networks DDCNet-B0 and DDCNet-B1}
       \label{fig:ErfB0vsB1}
    \end{figure}

    Figure~\ref{fig:ErfB0vsB1} shows the ERF of our proposed DDCNet models along with the extents of their FWHM. B0 has a smaller ERF extent that may limit its ability to estimate large motion such as in Sintel image sequences. This smaller but smoother ERF helped the network to perform well when estimating fine motions in the Middlebury dataset (See Table~\ref{tab:QuantitativeBenchmark}: AEE of 0.67 for B0 versus 1.2 for B1). 
    
    \subsubsection{Model Size / Compactness and Processing Time}
        For comparing the compactness of our DDCNet models with the lightweight and more elaborate models, we present Table~\ref{tab:RunTimeAndModelSize} with a summary of model parameters (number of layers, and number of learnable parameters) and computational speed of processing Sintel image sequences. 

        \begin{table}
            \centering
            \caption[Number of trainable parameters and computational of models]{Number of trainable parameters and computational speed of processing Sintel image sequences for lightweight and heavyweight models. Runtime is measured using Sintel image sequences with a frame size of \(1024 \times 436\) pixels. General speed differences between the computational frameworks such as Tensorflow and Caffe should be considered when comparing the run times of various networks.}
            \scalebox{0.6}{\begin{tabular}{ l|l|l|l|l|l|l}
			\toprule
			\tabhead{Method} & \tabhead{Number of} & \tabhead{Number of} &  \tabhead{Framework} & \tabhead{GPU (NVIDIA)} & \tabhead{Time (ms)} & \tabhead{FPS} \\
			 & \tabhead{layers} & \tabhead{parameters (m)} & & & & \\
			\midrule
			\tabhead{DDCNet-B0} & 31 & 1.03 & TF2 &  Quadro RTX 8000 & 76 & 13\\
			\tabhead{DDCNet-B1}  & 30 & 2.99 & TF2 &  Quadro RTX 8000 & 30 & 33\\
			  &  &  & \textit{Possible Caffe} &  Quadro RTX 8000 & \textit{\(\approx\)7} & \textit{142}\\
			\hline\\
			\tabhead{FlowNet Simple} & 17 & 38 & TF1 &   Tesla K80 & 86 & 11\\
			 &  &  & Caffe &     GTX  1080 & 18 & 55\\
			\tabhead{FlowNet Correlation} & 26 & 39.16 & TF1 &   Tesla K80 & 179 & 5\\
			 &  &  & Caffe &    GTX 1080 & 32 & 31\\
			\tabhead{FlowNet2} & 115 & 162.49 & TF1 &   Tesla K80 & 692 & 1\\ 
			 &  &  & Caffe &    GTX 1080 & 123 & 8\\ 
			\hline\\
			\tabhead{LiteFlowNet} & 94 & 5.37 & Caffe &    GTX  1080 & 88.53 & 12\\
			\tabhead{SPyNet} & 35 & 1.2 & Torch &    GTX  1080 & 129.83 & 8 \\
			\tabhead{PWC-Net+} & 59 & 8.75 & Caffe &  TITAN Xp & 39.63 & 25\\
			\bottomrule
\end{tabular}}
            \label{tab:RunTimeAndModelSize}
        \end{table}
        
        The number of trainable parameters for DDCNet-B0 and DDCNet-B1 was 1.03 and 2.99 million respectively which were comparable with the current compact / lightweight optical flow models.  In contrast, more elaborate and heavyweight models such as FlowNet models have 38 million to 162 million trainable parameters.
        
        All of our DDCNets were developed using Tensorflow. The existing lightweight models are mainly implemented using Caffe. FlowNet implementations are available in both Tensorflow and Caffe. In terms of processing speed, Caffe implementations are up to several times faster than Tensorflow models. Since our GPU has more RAM, we limit the batch size to 1 to make test times comparable.  With 30 ms processing time in Tensorflow, our DDCNet-B1 model is about 3 times faster than Tensorflow implementation of FlowNet-Simple on a less powerful GPU. B1 is also 3 times faster than LiteFlowNet implemented using Caffe and runs on GTX 1080.
        
        \begin{figure*}[htbp]
            \centering
            \scalebox{0.8}{\input{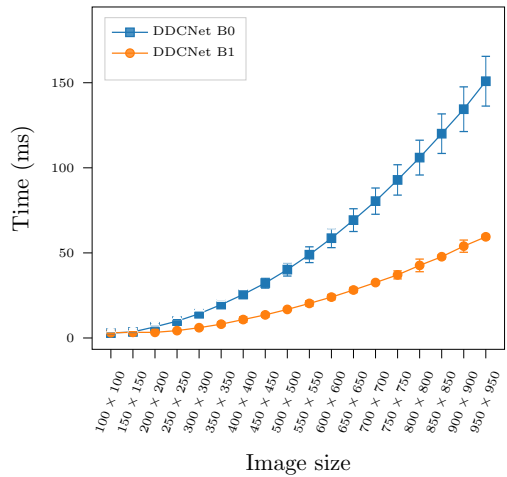}}
            \decoRule
            \caption[esting time of the DDCNet models as a function of sequence frame size]{Testing time of the DDCNet models as a function of sequence frame size.  B1 is computationally more efficient than B0 while being more accurate. Tests are performed on a single GPU Quadro RTX 8000 with batch sizes of 1.}
            \label{fig:TimeB0vsB1}
        \end{figure*}
        
        Figure~\ref{fig:TimeB0vsB1} shows the frame processing rate of both DDCNet models as a function of frame sizes. All DDCNet models are very fast (processing time per frame \(< 10\) ms) for frame size smaller than \(200 \times 200 \) pixels. B1 is the fastest model across all frame sizes. Though B0 has the least number of learnable parameters with full resolution feature maps within the network, the number of convolution operations is much higher compared to the B1 model.

    \subsubsection{Qualitative Results}
\begin{sidewaysfigure}
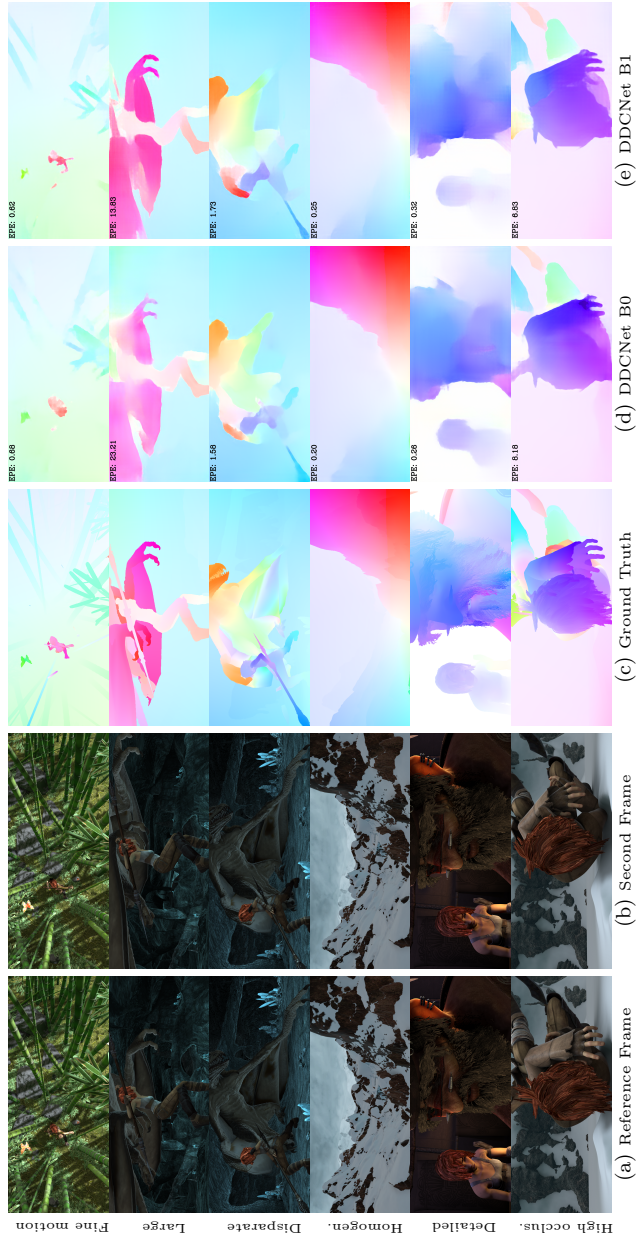

          \centering
          \scalebox{0.98}{\def\scale{0.166}
\def\rootDir{Figures/Qualitative/DDCNets}
\newcommand{\rott}[1]{\fontsize{1}{1}\selectfont{\rotatebox[origin=c]{90}{#1}}}

\newcommand{\IMG}[1]{\includegraphics[align=c,width=\scale\textwidth]{\rootDir/#1}}
\newcommand{\rowImg}[2]{
\xdef\id{#2}
\begin{minipage}[b]{0.01\linewidth}\centering \rott{#1} \end{minipage} & \IMG{GroundTruth/\id-f.png} & \IMG{GroundTruth/\id-s.png} & \IMG{GroundTruth/\id-gt.png} & \IMG{B0/\id-pr.png} & \IMG{B1/\id-pr.png}
}

\newcommand{\capNew}[1]{\begin{minipage}[b]{\scale\linewidth}\centering\subcaption{\tiny #1}\end{minipage}}
\setlength\tabcolsep{1.5pt}
\begin{tabular}{cccccc}
    \rowImg{Fine motion}{152}\\
    \rowImg{Large}{243}\\
    \rowImg{Disparate}{270}\\
    \rowImg{Homogen.}{364}\\
    \rowImg{Detailed}{391}\\
    \rowImg{High occlus.}{64}\\
    & \capNew{Reference Frame} & \capNew{Second Frame} & \capNew{Ground Truth} & \capNew{DDCNet B0} & \capNew{DDCNet B1}
\end{tabular}
}
          \decoRule
          \caption[Qualitative results: DDCNet-B0 vs B1]{Optical flow estimates of DDCNet-B0 vs DDCNet-B1 for qualitative assessment of the network performance}%
          \label{fig:QualitativeB0vsB1}%
        \end{sidewaysfigure}

        Figure~\ref{fig:QualitativeB0vsB1} shows performance of each of the DDCNet models on a broad variety of six example sequences with varied scene composition and motion dynamics from the \textit{Sintel Training Clean datasets}. The examples are namely 1) \textit{fine motion} sequence (first row in Figure~\ref{fig:QualitativeB0vsB1}) involving finer motions of smaller objects in various directions; 2) \textit{large motion} sequence (second row) involving higher flow velocities; 3) \textit{disparate motion} sequence (third row) involving varying flow dynamics and flow velocities in various scene segments; 4) \textit{homogeneous texture} sequence (fourth row) with motions involving scene segments with homogeneous texture; 5) \textit{high texture} sequence (fifth row) with motions involving highly textured scene segments; and 6) \textit{high occlusion} sequence (sixth row) with aperture problem including views of entire or partial object entering and / or leaving the scene.  
        
\begin{sidewaysfigure}
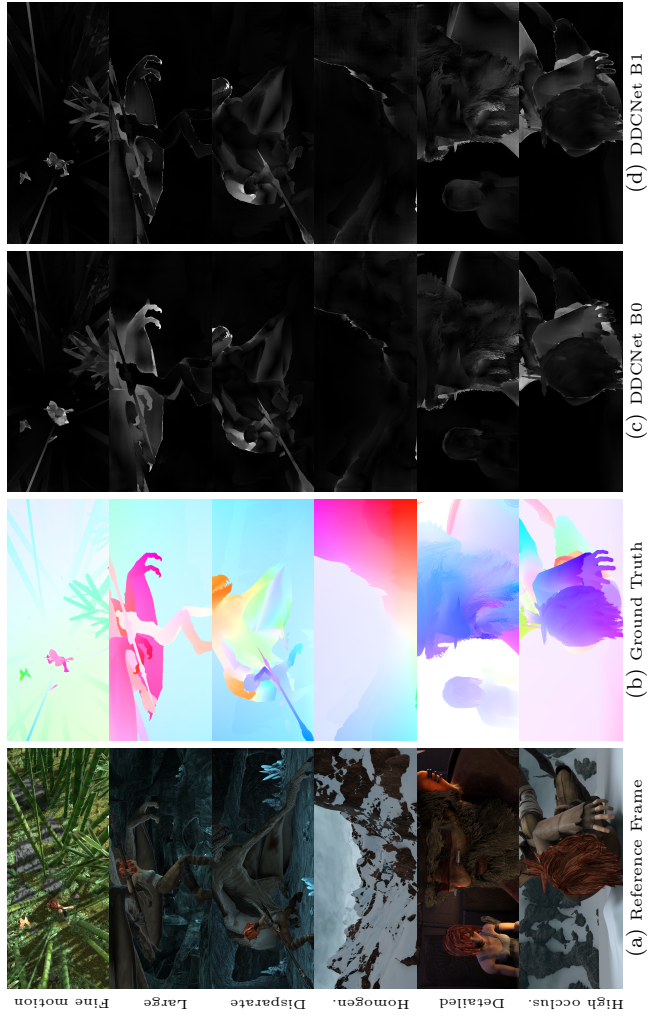

          \centering
          \scalebox{1.0}{\def\scale{0.166}
\def\rootDir{Figures/Qualitative/DDCNets}
\newcommand{\rott}[1]{\fontsize{1}{1}\selectfont{\rotatebox[origin=c]{90}{#1}}}

\newcommand{\IMG}[1]{\includegraphics[align=c,width=\scale\textwidth]{\rootDir/#1}}

\newcommand{\rowImg}[2]{
\xdef\id{#2}
\begin{minipage}[b]{0.01\linewidth}\centering \rott{#1} \end{minipage} & \IMG{GroundTruth/\id-f.png} & \IMG{GroundTruth/\id-gt.png} & \IMG{B0/\id-dis.png} & \IMG{B1/\id-dis.png}}

\newcommand{\capNew}[1]{\begin{minipage}[b]{\scale\linewidth}\centering\subcaption{\tiny #1}\end{minipage}}
\setlength\tabcolsep{1.5pt}
\begin{tabular}{ccccc}
    \rowImg{Fine motion}{152}\\
    \rowImg{Large}{243}\\
    \rowImg{Disparate}{270}\\
    \rowImg{Homogen.}{364}\\
    \rowImg{Detailed}{391}\\
    \rowImg{High occlus.}{64}\\
    & \capNew{Reference Frame} & \capNew{Ground Truth} & \capNew{DDCNet B0} & \capNew{DDCNet B1}
\end{tabular}
}
          \decoRule
          \caption[Error maps for DDCNet methods on selected examples]{Error maps of DDCNet-B0 vs B1 based on optical flow estimates shown in Figure~\ref{fig:QualitativeB0vsB1} for quantitative assessment of the network performance}%
          \label{fig:QualitativeB0vsB1Error}%
        \end{sidewaysfigure}

        Figure~\ref{fig:QualitativeB0vsB1Error} shows the corresponding error maps for each of the example sequences. The error map represents the magnitude of the difference between the estimated and ground-truth flow vector at each pixel.  Therefore, locations with larger estimation errors will appear brighter and locations with lower estimation errors will appear darker in the error maps.

         Difficulties of the B0 model in detecting larger motions and providing clear motion boundaries in the \textit{larger motion} sequence (e.g. near the dragon's legs in the scene) are likely due to the smaller extent of its ERF. With a broader ERF extent, B1 model was able to detect larger motions and with significantly clearer motion boundaries.  In image sequences with multi-directional flows in the scene (evident from multiple colors in the color-coded optical flow) as in the \textit{disparate motion} sequence, we observed that activations of neurons in the latest layers due to multi-directional flows could cancel each other leading to less accurate flow estimates.  The performance of B0 is better than B1 on this sequence. DDCNet models performed well on \textit{homogeneous texture} sequence. For the \textit{high texture} sequence both methods have difficulties identifying fine-grained flow estimates and motion boundaries in the high-textured zones. B0 was able to estimate finer flow velocities compared to B1.  For the most challenging \textit{high occlusion} sequence, both models faced difficulties in the occluded zones or zones with aperture problem.

 \subsection{Quantitative Evaluation on Benchmark Datasets}  
    Quantitative performance of our DDCNet models and other similar lightweight and heavyweight models for the Sintel, KITTI12, KITTI15, and Middlebury datasets are presented in Table~\ref{tab:QuantitativeBenchmark}. DDCNet-B1 performed better than B0 on image sequences with larger motions as in the Sintel dataset.  For datasets with large displacements such as Sintel, almost always B1 outperformed B0. For example, B0 failed in regions with motion larger than 250 pixels in the `large motion' sequence (second row) in Figure~\ref{fig:QualitativeB0vsB1Error} and in Figure~\ref{fig:QualitativeB0vsB1}. But for Middlebury with small to moderate flow velocities, B0 performed better than B1. 
    \begin{table}[h]
       \centering
       \caption[Quantitative results of different methods on benchmark datasets]{Average endpoint error of different methods on benchmark datasets. Entries with parentheses indicate the testing performance on data that was previous used for fine-tuning the network. $Fl_{\text{all}}$ measures the percentage of outlier estimates; pixels with EE $\ge 3$  and EE $\ge 5\%$ of the magnitude of its ground-truth flow vector were counted as outliers.
       } 
       \scalebox{0.8}{\scalebox{0.70}{
\begin{tabular}{|c|l||c c|c c|c c|c c c|c c|}
            \hline
\multirow{1}{*}{} 
&\multirow{1}{*}{\tabhead{Method}}   	                             	
&\multicolumn{2}{c|}{\tabhead{Sintel clean}}				
&\multicolumn{2}{c||}{\tabhead{Sintel final}}						
&\multicolumn{2}{c|}{\tabhead{KITTI12}}
&\multicolumn{3}{c||}{\tabhead{KITTI15}}   
&\multicolumn{2}{c|}{\tabhead{Middlebury}}\\

\multirow{1}{*}{}
&\multirow{1}{*}{}
&\multicolumn{1}{c}{train}&\multicolumn{1}{c|}{test}
&\multicolumn{1}{c}{train}&\multicolumn{1}{c||}{test}
&\multicolumn{1}{c}{train}&\multicolumn{1}{c|}{test}
&\multicolumn{1}{c}{train}&\multicolumn{1}{c}{train}&\multicolumn{1}{c||}{test}		
&\multicolumn{1}{c}{train}&\multicolumn{1}{c|}{test}\\	

\multirow{1}{*}{}
&\multirow{1}{*}{}
&\multicolumn{1}{c}{}&\multicolumn{1}{c|}{}
&\multicolumn{1}{c}{}&\multicolumn{1}{c||}{}
&\multicolumn{1}{c}{}&\multicolumn{1}{c|}{}
&\multicolumn{1}{c}{}&\multicolumn{1}{c}{(Fl-all)}&\multicolumn{1}{c||}{ (Fl-all)}		
&\multicolumn{1}{c}{}&\multicolumn{1}{c|}{}\\

\hline\hline    
\multirow{7}{*}{\rotatebox[origin=c]{90}{\tabhead{Heavyweight CNN}}}
&\multirow{1}{*}{FlowNetS~\cite{dosovitskiy2015flownet}}				
&4.50&\multicolumn{1}{c|}{7.42}	           
&5.45&\multicolumn{1}{c||}{8.43}
&8.26&\multicolumn{1}{c|}{}
& &\multicolumn{1}{c}{}&\multicolumn{1}{c||}{}			
&1.09&\multicolumn{1}{c|}{}\\       

\multirow{1}{*}{}
&\multirow{1}{*}{FlowNetS ft-sintel~\cite{dosovitskiy2015flownet}}				
&(3.66)&\multicolumn{1}{c|}{6.96}	           
&(4.44)&\multicolumn{1}{c||}{7.76}
&7.52&\multicolumn{1}{c|}{9.1}
& &\multicolumn{1}{c}{}&\multicolumn{1}{c||}{}			
&0.98&\multicolumn{1}{c|}{} \\ 
                                 
\multirow{1}{*}{}
&\multirow{1}{*}{FlowNetC~\cite{dosovitskiy2015flownet}}				
&4.31&\multicolumn{1}{c|}{7.28}	           
&5.87&\multicolumn{1}{c||}{8.81}
&9.35&\multicolumn{1}{c|}{}	
& &\multicolumn{1}{c}{}&\multicolumn{1}{c||}{}	
&1.15&\multicolumn{1}{c|}{}\\

\multirow{1}{*}{}
&\multirow{1}{*}{FlowNetC ft-sintel~\cite{dosovitskiy2015flownet}}				
&(3.78)&\multicolumn{1}{c|}{6.85}	           
&(5.28)&\multicolumn{1}{c||}{8.51}
&8.79&\multicolumn{1}{c|}{}	
& &\multicolumn{1}{c}{}&\multicolumn{1}{c||}{}		
&0.93&\multicolumn{1}{c|}{} \\
  



\multirow{1}{*}{}
&\multirow{1}{*}{FlowNet2~ \cite{ilg2017flownet}}				
&{2.02}&\multicolumn{1}{c|}{{3.96}}	           
&{3.54}&\multicolumn{1}{c||}{6.02}
&4.01&\multicolumn{1}{c|}{}
&10.08&\multicolumn{1}{c}{29.99\%}&\multicolumn{1}{c||}{}			
&{0.35}&\multicolumn{1}{c|}{{0.52}} \\ 

\multirow{1}{*}{}                                                                                            
&\multirow{1}{*}{FlowNet2 ft-sintel~\cite{ilg2017flownet}}				
&(1.45)&\multicolumn{1}{c|}{4.16}	           
&(2.19)&\multicolumn{1}{c||}{{5.74}}
&{3.54}&\multicolumn{1}{c|}{}
&{9.94}&\multicolumn{1}{c}{{28.02\%}}&\multicolumn{1}{c||}{}			
&{0.35}&\multicolumn{1}{c|}{} \\ 

\multirow{1}{*}{}
&\multirow{1}{*}{FlowNet2 ft-kitti~\cite{ilg2017flownet}}				
&3.43&\multicolumn{1}{c|}{}	           
&4.83&\multicolumn{1}{c||}{}
&(1.43)&\multicolumn{1}{c|}{{1.8}}
&(2.36)&\multicolumn{1}{c}{(8.88\%)}&\multicolumn{1}{c||}{{11.48\%}}			
&0.56&\multicolumn{1}{c|}{} \\
\multirow{1}{*}{}
&\multirow{1}{*}{}				
&&\multicolumn{1}{c|}{}	           
&&\multicolumn{1}{c||}{}
&&\multicolumn{1}{c|}{{}}
&&\multicolumn{1}{c}{}&\multicolumn{1}{c||}{{}}			
&&\multicolumn{1}{c|}{} \\

\hline\hline        
\multirow{12}{*}{\rotatebox[origin=c]{90}{\tabhead{Lightweight CNN}}}
\multirow{1}{*}{}
&\multirow{1}{*}{SPyNet~\cite{ranjan2017optical}}				
&4.12&\multicolumn{1}{c|}{6.69}	           
&5.57&\multicolumn{1}{c||}{8.43}
&9.12&\multicolumn{1}{c|}{}
& &\multicolumn{1}{c}{}&\multicolumn{1}{c||}{}		
&{0.33}&\multicolumn{1}{c|}{{0.58}} \\  

\multirow{1}{*}{}
&\multirow{1}{*}{SPyNet ft-sintel~\cite{ranjan2017optical}}				
&(3.17)&\multicolumn{1}{c|}{6.64}	           
&(4.32)&\multicolumn{1}{c||}{8.36}
&{3.36}&\multicolumn{1}{c|}{4.1}	
& &\multicolumn{1}{c}{}&\multicolumn{1}{c||}{35.07\%}	
&0.33&\multicolumn{1}{c|}{0.58} \\  

\multirow{1}{*}{}
&\multirow{1}{*}{PWC-Net+ ft-sintel~\cite{sun2019models}}				
&(1.71)&\multicolumn{1}{c|}{3.45}	           
&(2.34)&\multicolumn{1}{c||}{4.60}
&&\multicolumn{1}{c|}{}	
& &\multicolumn{1}{c}{}&\multicolumn{1}{c||}{}	
&&\multicolumn{1}{c|}{} \\

\multirow{1}{*}{}
&\multirow{1}{*}{PWC-Net+ ft-kitti~\cite{sun2019models}}				
&&\multicolumn{1}{c|}{}	           
&&\multicolumn{1}{c||}{}
&(0.99)&\multicolumn{1}{c|}{1.4}	
&(1.47)&\multicolumn{1}{c}{(7.59\%)}&\multicolumn{1}{c||}{7.72\%}	
& &\multicolumn{1}{c|}{} \\

\multirow{1}{*}{}
&\multirow{1}{*}{LiteFlowNet~\cite{hui2018liteflownet}}				
&{2.48}&\multicolumn{1}{c|}{}	           
&{4.04}&\multicolumn{1}{c||}{}
&{4.00}&\multicolumn{1}{c|}{}		
&{10.39}&\multicolumn{1}{c}{{28.50\%}}&\multicolumn{1}{c||}{}		
&0.39&\multicolumn{1}{c|}{} \\   

\multirow{1}{*}{}
&\multirow{1}{*}{LiteFlowNet ft-sintel~\cite{hui2018liteflownet}}				
&(1.64)&\multicolumn{1}{c|}{{4.86}}	           
&(2.23)&\multicolumn{1}{c||}{6.09}
&&\multicolumn{1}{c|}{}		
&&\multicolumn{1}{c}{}&\multicolumn{1}{c||}{}		
&&\multicolumn{1}{c|}{} \\

\multirow{1}{*}{}
&\multirow{1}{*}{LiteFlowNet ft-kitti~\cite{hui2018liteflownet}}				
&&\multicolumn{1}{c|}{}	           
&&\multicolumn{1}{c||}{}
&(1.29)&\multicolumn{1}{c|}{{1.7}}		
&(2.16)&\multicolumn{1}{c}{(8.16\%)}&\multicolumn{1}{c||}{{10.24\%}}		
&&\multicolumn{1}{c|}{} \\   

\multirow{1}{*}{}
&\multirow{1}{*}{LiteFlowNet2 ft-sintel~\cite{hui2019lightweight}}				
&(1.30)&\multicolumn{1}{c|}{3.48}	           
&(1.62)&\multicolumn{1}{c||}{4.69}
&&\multicolumn{1}{c|}{}	
& &\multicolumn{1}{c}{}&\multicolumn{1}{c||}{}	
&&\multicolumn{1}{c|}{} \\

\multirow{1}{*}{}
&\multirow{1}{*}{LiteFlowNet2 ft-kitti~\cite{hui2019lightweight}}				
&&\multicolumn{1}{c|}{}	           
&&\multicolumn{1}{c||}{}
&(0.95)&\multicolumn{1}{c|}{1.4}	
&(1.33)&\multicolumn{1}{c}{(4.32\%)}&\multicolumn{1}{c||}{7.62\%}	
&&\multicolumn{1}{c|}{} \\

\multirow{1}{*}{}
&\multirow{1}{*}{LiteFlowNet3 ft-sintel~\cite{hui2020liteflownet3}}				
&(1.32)&\multicolumn{1}{c|}{2.99}	           
&(1.76)&\multicolumn{1}{c||}{4.45}
&&\multicolumn{1}{c|}{}	
& &\multicolumn{1}{c}{}&\multicolumn{1}{c||}{}	
&&\multicolumn{1}{c|}{} \\

\multirow{1}{*}{}
&\multirow{1}{*}{LiteFlowNet3 ft-kitti~\cite{hui2020liteflownet3}}				
&&\multicolumn{1}{c|}{}	           
&&\multicolumn{1}{c||}{}
&(0.91)&\multicolumn{1}{c|}{1.3}	
&(1.26)&\multicolumn{1}{c}{(3.82\%)}&\multicolumn{1}{c||}{7.34\%}	
&&\multicolumn{1}{c|}{} \\

\cline {2-13}

\multirow{1}{*}{}
&\multirow{1}{*}{DDCNet-B0  ft-sintel}				
&(2.71)&\multicolumn{1}{c|}{7.20}           
&(3.27)&\multicolumn{1}{c||}{7.46}
&7.35&\multicolumn{1}{c|}{}		
&15.29&\multicolumn{1}{c}{47.78\%}&\multicolumn{1}{c||}{{}}		
&0.67&\multicolumn{1}{c|}{} \\    

\multirow{1}{*}{}
&\multirow{1}{*}{DDCNet-B1}				
&4.12&\multicolumn{1}{c|}{}           
&5.46&\multicolumn{1}{c||}{}
&9.57&\multicolumn{1}{c|}{}		
&16.43&\multicolumn{1}{c}{59.03\%}&\multicolumn{1}{c||}{}
&1.2&\multicolumn{1}{c|}{} \\  

\multirow{1}{*}{}
&\multirow{1}{*}{DDCNet-B1  ft-sintel}				
&(1.96)&\multicolumn{1}{c|}{6.19}           
&(2.25)&\multicolumn{1}{c||}{6.91}
&6.65&\multicolumn{1}{c|}{}		
&13.22&\multicolumn{1}{c}{52.68\%}&\multicolumn{1}{c||}{{}}		
&1.14&\multicolumn{1}{c|}{} \\  

\multirow{1}{*}{}
&\multirow{1}{*}{DDCNet-B1  ft-kitti}				
&6.65&\multicolumn{1}{c|}{}           
&8.38&\multicolumn{1}{c||}{}
&(1.76)&\multicolumn{1}{c|}{4.2}		
&(2.57)&\multicolumn{1}{c}{(15.56)\%}&\multicolumn{1}{c||}{38.23}	
&1.74&\multicolumn{1}{c|}{} \\  

\ifbone
\else
    \multirow{1}{*}{}
    &\multirow{1}{*}{DDCNet-Multires}			
    &2.71&\multicolumn{1}{c|}{}           
    &4.14&\multicolumn{1}{c||}{}
    &5.95&\multicolumn{1}{c|}{}		
    &13.54&\multicolumn{1}{c}{43.12\%}&\multicolumn{1}{c||}{}	
    &0.49&\multicolumn{1}{c|}{} \\
    
    \multirow{1}{*}{}
    &\multirow{1}{*}{DDCNet-Multires ft-sintel}			
    &{(1.36)}&\multicolumn{1}{c|}{5.34}           
    &{(1.70)}&\multicolumn{1}{c||}{{5.86}}
    &5.41&\multicolumn{1}{c|}{}		
    &12.59&\multicolumn{1}{c}{40.30\%}&\multicolumn{1}{c||}{}		
    &0.58&\multicolumn{1}{c|}{} \\
    
    \multirow{1}{*}{}
    &\multirow{1}{*}{DDCNet-Multires ft-kitti}			
    &6.86&\multicolumn{1}{c|}{}           
    &7.54&\multicolumn{1}{c||}{}
    &{(0.92)}&\multicolumn{1}{c|}{{3.2}}		
    &{(1.33)}&\multicolumn{1}{c}{{(5.59\%)}}&\multicolumn{1}{c||}{24.66\%}		
    &0.72&\multicolumn{1}{c|}{} \\
\fi

\hline
\end{tabular}
}}
       \label{tab:QuantitativeBenchmark}
    \end{table}
  
\begin{sidewaysfigure}
      \centering
      \scalebox{0.98}{\def\scale{0.166}
\def\rootDir{Figures/Qualitative/Sintel}
\newcommand{\rott}[1]{\fontsize{1}{1}\selectfont{\rotatebox[origin=c]{90}{#1}}}

\newcommand{\image}[2]{\begin{tikzpicture}[baseline=(current bounding box.center)]%
    \node[anchor=north west, inner sep=0, outer sep=0] (image) at (0,0){\includegraphics[align=c,width=\scale\textwidth]{\rootDir/#1}};%
    \def\temp{#2}\ifx\temp\empty%
    \else
       \node[lossLableStyle](loss) at (0.1,-0.1) {EPE: #2};%
    \fi%
\end{tikzpicture}}%

\newcommand{\rowImg}[6]{
\xdef\id{#2}
\begin{minipage}[b]{0.01\linewidth}\centering \rott{#1}\end{minipage} & \image{Frames/\id.png}{} & \image{GroundTruth/\id.png}{} & \image{FlowNetS/Flow/\id.png}{#3} & \image{FlowNet2/Flow/\id.png}{#4} & \image{LiteFlowNet3/Flow/\id.png}{#5} & \ifbone \image{DDCNetB1/Flow/\id.png}{#6} \else \image{DDCNetMultires/Flow/\id.png}{#6} \fi
}

\newcommand{\capNew}[1]{\begin{minipage}[b]{\scale\linewidth}\centering\subcaption{\tiny #1}\end{minipage}}
\setlength\tabcolsep{1.5pt}

\begin{tabular}{ccccccc}
    \rowImg{Market 3}{01}{1.66}{1.32}{1.10}{\ifbone 2.28 \else 1.62 \fi}\\
    \rowImg{Shaman 1}{02}{1.09}{0.79}{0.46}{\ifbone 1.65 \else 0.89 \fi}\\
    \rowImg{Ambush 1}{03}{37.30}{37.50}{33.17}{\ifbone 35.50 \else 30.73 \fi}\\
    \rowImg{Ambush 3}{04}{10.47}{6.97}{4.85}{\ifbone 7.60 \else 7.12 \fi}\\
    \rowImg{Bamboo 3}{05}{1.22}{1.02}{0.96}{\ifbone 1.99 \else 1.37 \fi}\\
    \rowImg{Cave 3}{06}{3.87}{5.45}{4.64}{\ifbone 8.27 \else 7.10 \fi}\\
    & \capNew{Reference Frame} & \capNew{Ground Truth} & \capNew{FlowNet Simple} & \capNew{FlowNet2} & \capNew{LiteFlowNet3} & \ifbone \capNew{DDCNet B1}
    \else \capNew{DDCNet Multires} \fi%
\end{tabular}

}
      \decoRule
      \caption[Qualitative comparison between our methods and others]{Performance of our DDCNet-B1, lightweight and heavyweight models for estimating optical flow for selected Sintel image sequences.}%
      \label{fig:QualitativeSintel}%
    \end{sidewaysfigure}
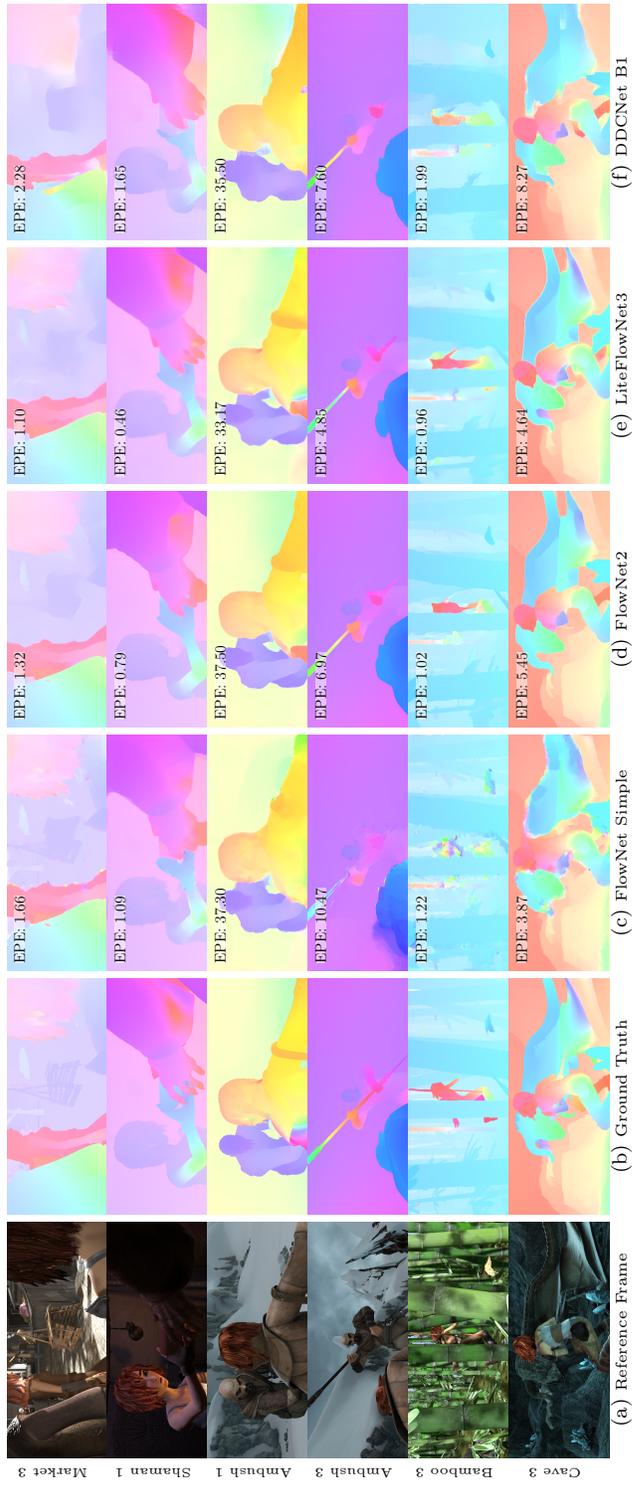

    As mentioned earlier, FlowNet-Simple was the first successful CNN-based model that was on par with the other state-of-the-art models \cite{Dosovitskiy2015}. Since then, it has been used as a building block for numerous optical flow estimation methods \cite{ilg2017flownet, jason2016back, ren2017unsupervised, wang2018occlusion}. The principal goal of this research work is to redesign this building block more efficiently and more effectively for optical flow estimation.  Among the lightweight CNN models, LiteFlowNet is one of the top-performing networks. Therefore, we focus on comparing the performance of our DDCNets with FlowNet and LiteFlowNet models. Figure~\ref{fig:QualitativeSintel} shows the ground truth optical flow and optical flow estimated by FlowNet-Simple, FlowNet2, LiteFlowNet, and DDCNet-B1.  As summarized in Table~\ref{tab:RunTimeAndModelSize}, FlowNet-Simple has 12 times more parameters (38 million vs 2.99) than DDCNet-B1. Despite having an extra variational-based refiner step on top of the main network, FlowNet-Simple has a higher endpoint error on almost all of the benchmark datasets. Superior quality of DDCNet-B1 estimates when compared to FlowNet-Simple can be observed in Figure~\ref{fig:QualitativeSintel}.
    
    The endpoint error of FlowNet2 is better than our DDCNet models, but FlowNet2 has 25 times more learnable parameters than our model. Further, FlowNet2 is more than 6 times slower during testing and it is expected to require a longer training duration since each network needs to be trained separately.

    In the lightweight class, DDCNet-B1 outperforms SpyNet \cite{ranjan2017optical}, but admittedly, LiteFlowNet models have better performance in terms of AEE but are slower. While these models have a comparable number of parameters with DDCNet models, they perform expensive feature matching operations. These models use custom layers which makes them perform well only on optical flow estimation whereas our models can be easily applied to other dense estimation tasks. Because we are not building any explicit spatial pyramid such as LiteFlowNet, our models do not suffer from the vanishing problem related to faster moving objects with smaller spatial extents in a scene that are often encountered in classical multi-resolution algorithms. This can be observed from fourth row (sequence \emph{Ambush 3}) of Figure~\ref{fig:QualitativeSintel}. A narrow, stick-like object that moves fast is apparent in this sample. DDCNet-B1 and FlowNet2 are able to estimate its motion, but LiteFlowNet has difficulties in that region.

\section{Software}
DDCNet models along with necessary instructions for running the software are available to the public in the following URL:\\ \href{https://github.com/alisaaalehi/DDCNet}{https://github.com/alisaaalehi/DDCNet}.

\section{Conclusion}

In this work, we present two key strategies or recommendations for building a successful dense prediction network namely 1) preserving spatial information throughout the network (preferably implicitly by keeping extracted features at each layer in their respective spatial order by avoiding spatial aggregation operations such as pooling) and 2) designing networks with large-enough receptive field (effective receptive field) for the output units to enhance their ability to provide an accurate estimate for the task (e.g. spatiotemporal matching). While these strategies are necessary, they are not sufficient. A generic design with a sufficiently larger receptive field and high-resolution features can be further refined and tailored to more specific computer vision tasks such as disparity estimation and semantic segmentation.  Through our design strategy, we have demonstrated that we can design CNNs to perform complex long-range matching tasks without any need for explicit matching layers in the network. The benchmark results (Sintel, KITTI, and Middlebury) indicate that our compact networks can achieve comparable performance in the class of lightweight optical flow networks. Further, such DDCNet models (sub-nets) can be used as building blocks in more complex network designs with a higher estimation accuracy (similar to the current usage of FlowNet-Simple as a building block in more complex network designs). The model itself can be scaled up by either the increasing number of sub-nets or by increasing the number of layers in each of the sub-nets for better performance and will be presented in a future study. 

\section*{Acknowledgement}
The research was supported in part by financial support in the form of a Herff Fellowship from the Herff College of Engineering, The University of Memphis (UoM); tuition fees support from the Department of Electrical and Computer Engineering, UoM; a summer student fellowship from the \emph{Fight for Sight} organization; and in part by the National Institutes of Health, National Eye Institute Grant EY020518.


\pagebreak
\bibliographystyle{plainnat}
\bibliography{References.bib}
\end{document}